# Plant Disease Detection through Multimodal Large Language Models and Convolutional Neural Networks

Konstantinos I. Roumeliotis, Ranjan Sapkota, Manoj Karkee, Nikolaos D. Tselikas, and Dimitrios K. Nasiopoulos

*Abstract*— Automation in agriculture plays a vital role in addressing challenges related to crop monitoring and disease management, particularly through early detection systems. This study investigates the effectiveness of combining multimodal Large Language Models (LLMs), specifically GPT-4o, with Convolutional Neural Networks (CNNs) for automated plant disease classification using leaf imagery. Leveraging the PlantVillage dataset, we systematically evaluate model performance across zero-shot, few-shot, and progressive fine-tuning scenarios. A comparative analysis between GPT-4o and the widely used ResNet-50 model was conducted across three resolutions (100, 150, and 256 pixels) and two plant species (apple and corn). Results indicate that fine-tuned GPT-4o models achieved slightly better performance compared to the performance of ResNet-50, achieving up to 98.12% classification accuracy on apple leaf images, compared to 96.88% achieved by ResNet-50, with improved generalization and near-zero training loss. However, zero-shot performance of GPT-4o was significantly lower, underscoring the need for minimal training. Additional evaluations on cross-resolution and cross-plant generalization revealed the models' adaptability and limitations when applied to new domains. The findings highlight the promise of integrating multimodal LLMs into automated disease detection pipelines, enhancing the scalability and intelligence of precision agriculture systems while reducing the dependence on large, labeled datasets and high-resolution sensor infrastructure.

*Note to Practitioners*— Inspired by the power and strengths of Large Language Models (LLMs), this work shows that integrating fine-tuned multimodal LLMs like GPT-4o with CNNs advances automation and robotics in plant disease detection. This approach enables highly accurate, scalable, and early diagnosis using minimal labeled data and lower-resolution images, supporting efficient and sustainable crop management in modern agricultural and robotic systems.

*Index Terms*: Multimodal Large Language Models, Plant Disease Detection, Plant Disease Classification, Multimodal Classification, Deep Learning in Agriculture, Automation

## I. Introduction

AUTOMATION in agriculture has emerged as a transformative approach to combating critical challenges such as labor issues, nutrient optimization, plant disease control, which significantly threaten crop yield and quality, sustainable farming, and global food security [1, 2].

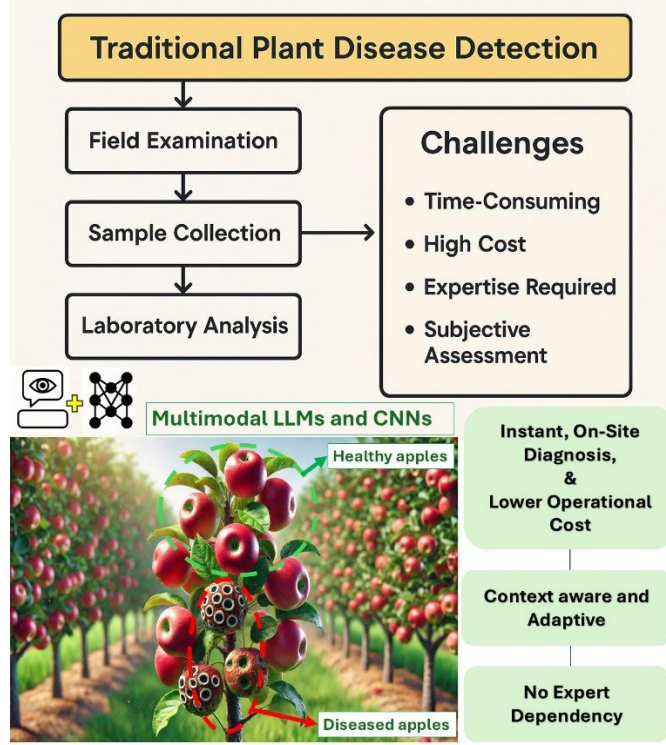

**Fig. 1.** Major challenges posed by plant diseases classification with traditional neural network-based approaches. The figure introduces the core problem addressed in this study. LLM + CNN based disease detection systems offer a promising solution for accurate plant disease classification and sustainable crop management.

Manuscript received -; accepted -. Date of publication -; date of current version -. This article was recommended for publication by Associate Editor - and Editor - upon evaluation of the reviewers' comments. This research was funded by National Institute of Food and Agriculture (US). The publication of the article in OA mode was financially supported by HEAL-Link. (Corresponding authors: Konstantinos I. Roumeliotis, Ranjan Sapkota, Manoj Karkee)

**Konstantinos I. Roumeliotis:** Department of Informatics and Telecommunications, University of the Peloponnese, 22131 Tripoli, Greece (e-mail: k.roumeliotis@uop.gr)

**Ranjan Sapkota**: Department of Biological & Environmental Engineering, Cornell University, Ithaca, 14850, NY, USA (e-mail: rs2672@cornell.edu)

**Manoj Karkee:** Department of Biological & Environmental Engineering, Cornell University, Ithaca, 14850, NY, USA (e-mail: mk2684@cornell.edu)

**Nikolaos D. Tselikas:** Department of Informatics and Telecommunications, University of the Peloponnese, 22131 Tripoli, Greece (e-mail: ntsel@uop.gr)

**Dimitrios K. Nasiopoulos**: Department of Agribusiness and Supply Chain Management, School of Applied Economics and Social Sciences, Agricultural University of Athens, 11855 Athens, Greece (e-mail: dimnas@aua.gr).

Given the increasing pressure of plant diseases on global agriculture, addressing this challenge has become more critical than ever. Disease outbreaks not only compromise crop health but also exacerbate the existing threats to food security, sustainable farming, and economic stability mentioned earlier. As shown in **Fig. 1**, diseased plants (e.g. apples) can lead to poor-quality harvests, reduced food production, and economic instability for farmers. To address these issues, automated crop monitoring leveraging intelligent sensing and diagnostic systems that enable early and accurate detection of plant diseases is essential, which will support timely interventions that limit damage and halt disease progression [3]. Early disease detection plays a key role in preventing widespread outbreaks and helps preserve both the quantity and quality of agricultural output [4, 5]. Automated monitoring systems also reduce dependence on manual labor, improve resource efficiency, and minimize economic losses resulting from delayed or inaccurate responses [6, 7].

Moreover, early disease detection contributes to cost efficiency in agricultural management. Detecting and addressing diseases at an early stage can drastically reduce the expenses associated with more aggressive treatments required at later stages [8]. For example, early intervention may require less intensive pesticide applications and lower labor costs compared to those necessary for controlling a full-blown disease outbreak [9, 10]. This early intervention is particularly important as it also facilitates the targeted use of pesticides. By precisely identifying affected areas, farmers can limit the application of chemicals to only necessary zones, thereby reducing the amount of chemical runoff into ecosystems and minimizing the environmental impact [11, 12].

Furthermore, early detection of plant diseases plays a crucial role in enhancing food security [13]. In regions heavily dependent on agriculture, ensuring the health of crops directly supports the stability of food supplies [14, 15]. This is vital for economic stability and for preventing hunger in vulnerable populations. Another significant benefit of early plant disease detection is the assurance of high-quality agricultural produce [16]. Products free from disease not only fetch higher market prices but also meet consumer expectations for quality, which is essential for maintaining trust and satisfaction among consumers [17, 18].

Recent advances in machine learning (ML) (e.g. deep learning (DL) models) have paved the way for innovative approaches in plant disease diagnosis [19]. These traditional neural networks based ands DL/ML based approaches for plant disease detection, though more efficient than manual methods, still face several critical limitations as illustrated in Fig. 1. These models typically rely heavily on single-modal image data and require extensive labeled datasets curated through labor-intensive field examination, sample collection, and laboratory analysis. The associated processes are time-consuming, costly, and demand significant domain expertise. Furthermore, their performance can be inconsistent under varying real-world conditions due to limited contextual understanding. Most conventional DL models also suffer from subjective biases and lack generalizability across diverse environments, restricting their scalability and reliability in real agricultural applications.

Integrating LLMs with CNNs enables real-time, on-site plant disease detection by combining visual analysis with contextual understanding from multimodal inputs like sensor data and text [1,2]. This fusion reduces operational costs, removes expert dependency, and delivers more accurate, objective, and adaptive diagnoses across diverse agricultural environments (**Fig. 1**). This study explores the potential of fine-tuning Generative Pre-trained Transformers (GPT) for multimodal classification tasks in agriculture, comparing their performance against conventional image-based models such as Residual Networks (ResNet). Utilizing the PlantVillage dataset, we conduct a comprehensive evaluation of these models across various scenarios, including zero-shot and few-shot learning at different image resolutions.

This study involves iterative fine-tuning cycles across different plant species and conditions to establish a robust framework for early disease detection in key crops such as apples and corn. The study initiates by selecting optimal hyperparameters for fine-tuning both GPT-4o and ResNet-50 models, using Bayesian optimization with 30 trials, early stopping, pruning, and a maximum of 10 epochs for ResNet-50, and subsequently hypothesizes that the same parameters may provide comparable performance when applied to the GPT-4o multimodal model across different plant types and image resolutions.

A few-shot fine-tuning procedure is subsequently applied to GPT-4o, using plant leaf image samples and their corresponding labels/diseases. The model is fine-tuned across six training and six validation sets, accounting for three resolutions (100, 150, and 256 pixels) and two plant types (apple and corn). Both models are then tasked with classifying plant leaf images in the test sets based on their respective diseases.

To validate our hypothesis regarding the transferability of hyperparameters optimized for ResNet-50, we implement a progressive fine-tuning approach for GPT-4o. Specifically, the model is iteratively fine-tuned with additional batches of 128 image samples in a staged training process. In contrast to the earlier hyperparameter settings, this phase employs default hyperparameters as recommended by OpenAI. The same four-phase progressive fine-tuning protocol is applied to both plant species across all resolutions. Following training, the fine-tuned models are evaluated on their respective test sets.

Given that GPT-4o is a pre-trained model capable of zero-shot classification, we also prompt the base GPT-4o model and its mini variant to classify the same test samples without fine-tuning. The study concludes with cross-resolution and cross-plant evaluations to assess the generalization capabilities of the best-performing models. Through this extensive investigation, we derive valuable insights into the suitability of multimodal LLMs for plant disease detection based on image classification.

In the next section (literature review), we present a brief discussion on plant disease detection methodologies, categorized into three distinct approaches: a) traditional methods employed prior to the advent of machine learning (ML), b) ML techniques, including deep learning (DL), that have revolutionized disease detection in agriculture, and c) the cutting-edge application of multimodal Large Language Models (LLMs) in this field. Each category will be explored to provide the capabilities and limitations of these techniques, which was then used to identify the research gap the proposed study will address.

## II. LITERATURE REVIEW

### A. Traditional Plant Disease Detection Methods

Before the advent of ML and DL, traditional image processing techniques have been used broadly to develop disease detection models. Techniques such as Color Histogram Analysis and Threshold-Based Segmentation utilized RGB and HSV color spaces to identify diseased areas through abnormal pigmentation analysis [20, 21]. However, these methods faced significant challenges, notably their high sensitivity to lighting variations and camera settings, which often resulted in the inability to distinguish diseases with similar color patterns [22, 23]. Similarly, Texture Analysis through Gray-Level Co-occurrence Matrices (GLCM) [24], color texture [25] and Edge Detection methods like Canny [26, 27] and Sobel [28, 29] were employed to quantify textural features and detect lesion boundaries. These techniques, while insightful, required high-resolution images and were particularly susceptible to noise, making them less effective in field conditions where overlapping textures and early-stage disease symptoms presented subtle cues [30].

In addition, Morphological Shape Analysis-based approaches were used to measure geometric properties of lesions to quantify disease severity [31], while Thermal Imaging and Fluorescence Imaging Spectroscopy were used to detect physiological changes caused by diseases [32, 33], such as temperature variations or chlorophyll degradation. Although these methods provided deeper insights into plant health, they required expensive, specialized equipment and were sensitive to environmental factors, making them less practical for field use [34]. Similarly, Stereoscopic 3D Imaging [35] and Multispectral/Hyperspectral [36] Imaging techniques offered advanced diagnostics by reconstructing 3D leaf models and capturing reflectance spectra across wavelengths, respectively. However, these techniques were computationally intensive and required significant expertise and resources for data collection and analysis, limiting their application to controlled settings [37, 38].

Moreover, the integration of manual feature extraction pipelines with Support Vector Machines (SVMs) marked an early attempt to combine multiple handcrafted features for disease classification [39]. However, this approach was labor-intensive and lacked generalizability to previously unseen diseases and conditions, limiting its effectiveness in real-world scenarios [40].

### B. Machine Learning and Deep Learning Based Plant Disease Detection Methods

Traditional ML techniques have played a foundational role in advancing plant disease detection by offering data-driven alternatives to manual image analysis. Approaches such as Support Vector Machines (SVMs) and Random Forests (RFs) have been widely used for classifying plant diseases based on extracted features like color, shape, and texture [70, 71]. These models are often employed in conjunction with feature extraction pipelines, which rely on hand-crafted descriptors derived from either traditional image processing techniques or from the early layers of neural networks. While effective in many cases, traditional ML models are constrained by their dependence on domain-specific feature engineering and their limited ability to generalize across diverse environmental conditions or disease manifestations.

To address these limitations, deep learning (DL) has emerged as a powerful extension of ML, introducing increased model depth and complexity to learn hierarchical feature representations directly from raw data. The integration of DL into plant disease detection has significantly improved diagnostic accuracy and efficiency, particularly with the advent of Convolutional Neural Networks (CNNs) [41]. CNNs automatically extract relevant features from images through stacked convolutional layers that capture spatial hierarchies, such as edges, textures, and disease-specific patterns [42]. Notable architectures adapted for plant disease classification include AlexNet [43–45], VGG [46–49], and ResNet [50–54], each offering progressively deeper networks to improve feature discrimination and robustness [41, 55]. Inception networks further enhance performance by incorporating multiple kernel sizes within a single layer, enabling multi-scale feature detection in complex agricultural imagery [56].

Transfer learning has also gained traction in plant pathology applications, allowing models pre-trained on large datasets like ImageNet to be fine-tuned on smaller, domain-specific datasets [57, 58]. This is especially valuable in agriculture, where labeled data is often limited [41]. In addition to classification, segmentation models such as U-Net [59, 60] and Mask R-CNN [61–65] have been deployed to localize disease symptoms with pixel-level accuracy, facilitating precise assessment of disease extent and severity. More recent developments include the use of Generative Adversarial Networks (GANs) for synthetic data generation in low-resource settings. GANs can create realistic diseased plant images to augment training datasets, improving model robustness and generalization [66, 67]. Autoencoders, particularly variational autoencoders, are increasingly used for unsupervised anomaly detection, identifying diseased patterns as deviations from healthy plant imagery without requiring exhaustive labeling [68, 69].

Despite their advancements, traditional ML/DL models primarily rely on visual cues alone and lack the capacity to incorporate contextual or descriptive information about plant diseases. These models often struggle in scenarios with ambiguous visual symptoms, visually similar diseases, or limited labeled datasets, and they lack interpretability and adaptability in real-world, dynamic environments. Furthermore, they cannot effectively utilize textual agronomic knowledge, such as expert descriptions or symptom narratives, which limit their diagnostic comprehensiveness. These constraints underscore the need for more holistic and semantically aware systems. This gap has led to the emergence of multimodal LLMs, which combine vision and language understanding to overcome these limitations and enable richer, context-driven plant disease detection.

*C. Multimodal Large Language Model-based Plant Disease Detection Methods*

Recent advancements in multimodal LLMs have opened new opportunities for plant disease detection, combining the power of visual and textual data analysis. This approach addresses limitations of traditional computer vision methods by incorporating rich contextual information and natural language descriptions. A pioneering work in this field is the "Snap and Diagnose" system, introduced by Wei et al. (2024) [72]. This multimodal retrieval system utilizes the PlantWild dataset, comprising over 18,000 images across 89 disease categories. The system employs a novel CLIP-based vision-language model that encodes both disease descriptions and images into a shared latent space, enabling cross-modal retrieval. Users can input either plant disease images or textual descriptions to retrieve visually similar images, suggesting potential diagnoses [72].

Building on this concept, Zhou et al. (2024) [73] proposed a model exploring semantic embedding methods for disease images and description text. Their approach demonstrated improved performance in crop disease identification by fusing advanced visual features with encoded contextual information from natural language modalities. Likewise, the PepperNet model, introduced in late 2024, addresses the challenges of detecting pepper diseases and pests in complex scenarios [73]. This model utilizes multimodal pepper diseases and pests object detection dataset (PDD), which includes diverse images and detailed natural language descriptions. PepperNet utilizes fine-grained multimodal attribute contrastive learning to differentiate subtle local variations between visually similar disease symptoms, thereby improving detection accuracy in challenging conditions such as symptom similarity and partial leaf occlusion.

In August 2024, researchers proposed an in-the-wild multimodal plant disease recognition dataset, featuring the largest number of disease classes to date, along with text-based descriptions [74]. This dataset aims to benchmark and advance multimodal approaches in plant disease recognition, addressing the limitations of controlled environment datasets. Building on this progress, the application of multimodal transformer models in intelligent agriculture has further expanded the capabilities of disease detection systems. These models can provide in-depth understanding and suggestions by analyzing agricultural text materials alongside visual data, offering a more comprehensive approach to disease diagnosis [75].

Recent implementations have also integrated large language models with agricultural knowledge graphs [76, 77]. These systems enhance disease detection accuracy through innovative graph attention mechanisms and optimized loss functions, leveraging the vast knowledge encoded in LLMs and structured agricultural data.

While prior research has demonstrated the effectiveness of traditional CNN-based models like ResNet-50 in plant disease classification tasks, there remains a significant gap in exploring the capabilities of LLMs such as GPT-4o in this domain. Most existing approaches rely heavily on supervised learning with extensive labeled datasets and do not leverage the zero-shot and few-shot reasoning abilities of state-of-the-art LLMs. This study addresses this gap by systematically evaluating GPT-4o's performance in both zero-shot and fine-tuned settings, comparing it with a well-established baseline model, and extending the evaluation across variations in image resolution and plant species. Additionally, it proposes the effective transfer of hyperparameters from CNN hyperparameter tuning to LLM fine-tuning, achieving improved results with reduced experimentation. By doing so, we aim to assess the model's robustness, generalization capacity, and potential as a flexible, multimodal tool in precision agriculture.

### III. METHODOLOGY

This section outlines the methodological framework used to assess the capabilities of the GPT-4o multimodal LLM in detecting and classifying plant leaf diseases through its computer vision capabilities. The methodology is systematically structured into five distinct phases, each addressing a fundamental aspect of the research process. The first phase involves GPT-4o zero-shot predictions. In the second phase, few-shot fine-tuning of both GPT-4o and ResNet-50 is performed. Following the identification of optimal hyperparameters through a thorough fine-tuning process of the ResNet-50 model, model predictions are generated. The third phase focuses on progressive fine-tuning of GPT-4o using the default hyperparameters provided by OpenAI, accompanied by subsequent predictions. The fourth phase entails generating cross-resolution predictions to evaluate model robustness across varying image qualities. Finally, the fifth phase involves cross-plant predictions, aimed at assessing the model's generalization capability across different plant species.

Initially, we provide a comprehensive description of the dataset, detailing its construction, composition, and preprocessing techniques implemented to enhance data integrity. Given the inherent challenge of class imbalance, we outline the strategies employed to mitigate biases and ensure robust model training. Subsequently, we present the model selection process, prompt engineering methodologies, and fine-tuning strategies, with particular emphasis on the role of few-shot learning in optimizing model performance. The fine-tuning procedures for both the GPT-4o and ResNet-50 models are documented, underscoring their adaptability across diverse plant disease categories. Furthermore, challenges encountered during the fine-tuning of GPT-4o necessitated a progressive fine-tuning approach to systematically identify and address performance limitations within the model. To facilitate reproducibility and future research, all datasets, training classes, and evaluation code developed in this study are publicly available on GitHub as open-source under the Apache-2.0 license [78]. The overall methodological framework used in this study is illustrated in **Fig. 2**.

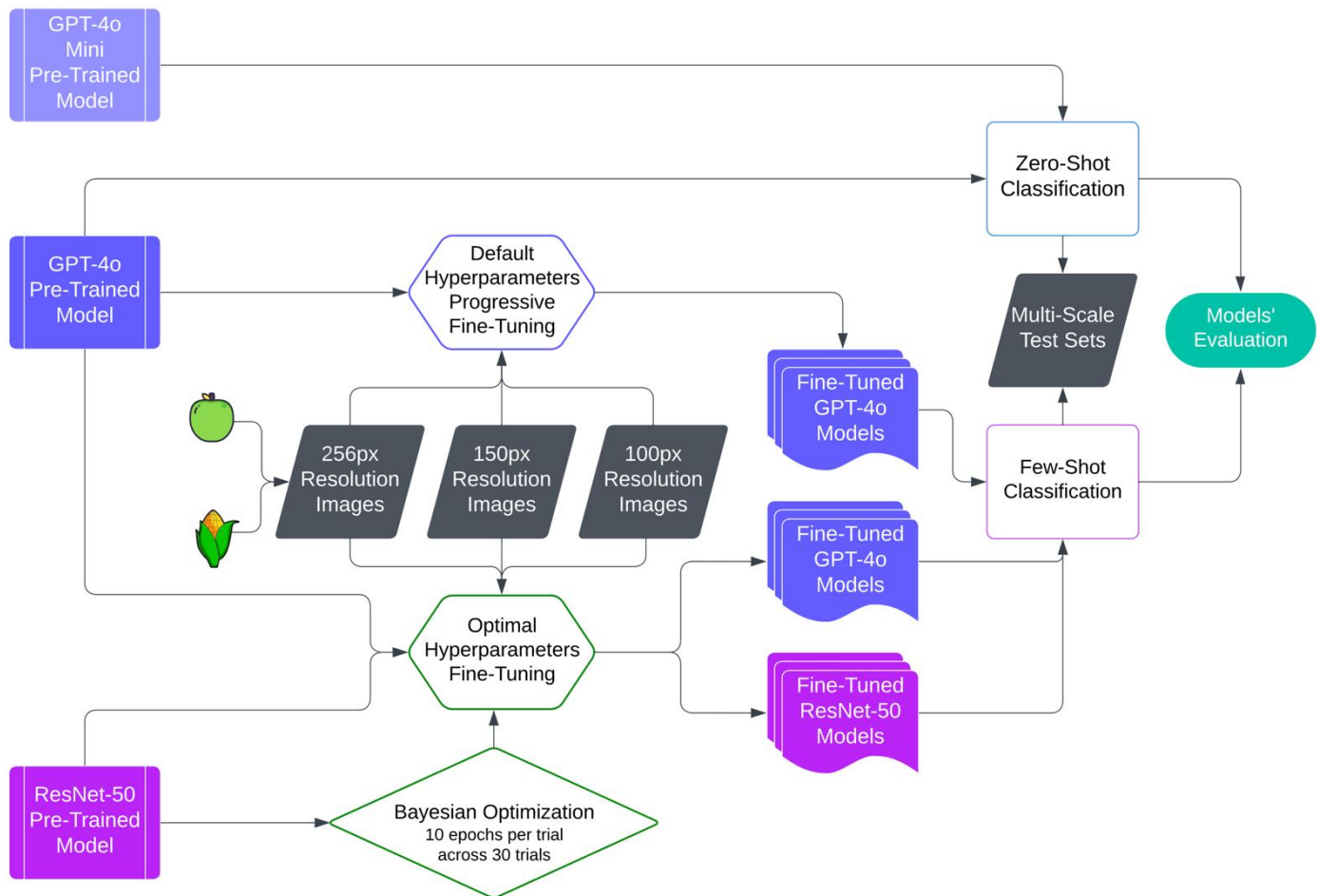

**Fig. 2.** Fine-Tuning and Evaluation Framework for Multimodal Plant Leaf Disease Classification Using GPT-4o and ResNet-50. The process begins with Bayesian optimization applied to ResNet-50 to determine optimal hyperparameters. These are used to fine-tune both ResNet-50 and GPT-4o models across varying image resolutions, followed by test set predictions. In parallel, GPT-4o is also progressively fine-tuned using OpenAI's default hyperparameters. Additionally, gpt-4o-mini and gpt-4o base models perform zero-shot predictions on the same test set. All models are then evaluated on predictive performance.

### A. Dataset

The dataset used in this study was sourced from the PlantVillage dataset, a widely cited and publicly available benchmark resource for plant disease detection tasks. As illustrated in **Fig. 3**, it contains a large collection of plant leaf images representing various disease conditions across multiple species, including apple, corn, blueberry, cherry, and grape [79]. Each image is manually annotated by the original dataset creators, ensuring high-quality labels aligned with specific plant disease categories [80]. The dataset is systematically organized into subfolders, with each folder corresponding to a distinct plant-disease class, thereby supporting supervised learning and class-specific analysis. Its popularity and structured format make it a reliable foundation for developing and evaluating plant disease classification models.

For this study, the following plant leaf samples were specifically selected:
- Apple leaves with the following labels:
    - Healthy (1,646 samples)
    - Scab (631 samples)
    - Black rot (622 samples)
    - Cedar rust (276 samples)
- Corn leaves with the following labels:
    - Healthy (1,163 samples)
    - Gray leaf spot (514 samples)
    - Northern leaf blight (986 samples)
    - Rust (1,193 samples)

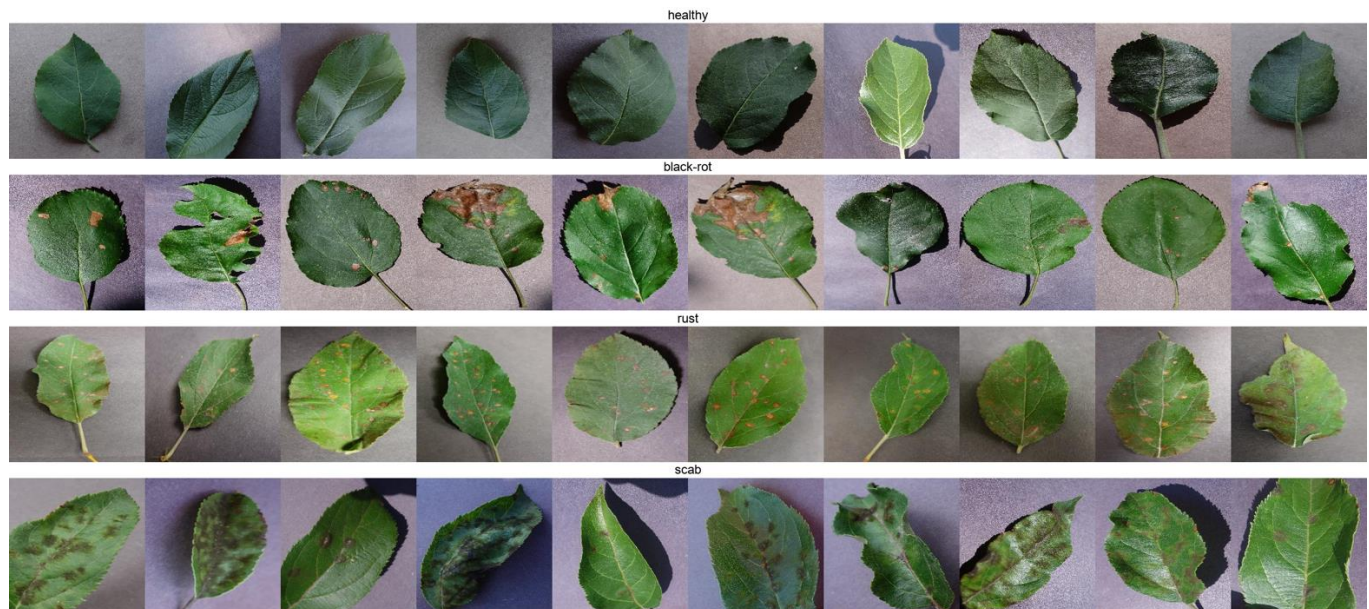
**Fig. 3.** Labeled images of apple leaves from the PlantVillage dataset, showing various plant diseases.

To ensure consistency, the selected images were downloaded while preserving the original folder structures, where each folder was named according to the corresponding plant disease. Each folder exclusively contained images of a specific disease for a given plant species.

A crucial step in fine-tuning and prediction involved constructing a structured dataset from these images. Using a Python script, we iterated through each folder, generating sample pairs consisting of an image and its corresponding disease label. These data pairs were subsequently stored in CSV format, resulting in two distinct datasets: one for apple leaf samples and another for corn leaf samples.

Through comprehensive validation, we confirmed that all images in the dataset maintained a uniform resolution of 256x256 pixels and were consistently formatted as JPEG files. This uniformity ensured compatibility across different model training and evaluation pipelines.

*B. Data Segmentation and Preprocessing*

There was an imbalance in disease/category distribution across samples of this dataset, which could potentially affect the fine-tuning process negatively. An imbalanced dataset often introduces biases toward majority classes, leading to model overfitting and poor generalization for minority classes. Various techniques can mitigate this issue, including oversampling, data augmentation, and undersampling. Due to the nature of our data (image-based), oversampling was not a viable option. Instead, data augmentation techniques—such as rotation, flipping, and cropping—could be employed to generate additional samples. However, to maintain balance across diseases and plant types, undersampling was selected as the primary approach. This method resulted in two balanced datasets: one for apple leaves and another for corn leaves, each containing 800 images distributed equally across plant diseases (200 images per disease).

**Investigating Image Size Impact**

A key aspect of this research is to explore how both the GPT-4o and ResNet-50 models process images of different resolutions and assess their predictive performance on cross-resolution datasets (the rationale for model selection is discussed in Section C). Deep convolutional neural networks, such as ResNet-50, are typically trained on images of approximately 224×224 pixels (e.g., ImageNet). Increasing image resolution, may introduce additional irrelevant information that the model is not trained to handle. If a model is not explicitly fine-tuned for larger images, it may struggle to adapt to the increased resolution, potentially degrading performance. However, larger images also contain more pixels and features, which could enhance generalization if properly fine-tuned.

To investigate these effects, different image sizes were generated after fine-tuning the models. Using the PIL library, thumbnail versions of each sample were created at 100×100 pixels, 150×150 pixels, and 256×256 pixels (original size), while preserving the aspect ratio. These images were organized into separate subfolders for better management.

Furthermore, all processed images were uploaded to a live web server. Rather than utilizing cloud storage solutions such as Google Cloud Buckets, we opted to host the images on our own web server to better manage file accessibility. While ResNet-50 can process locally stored images, GPT-4o, as an API-based closed-source model, requires images to be accessible via live URLs for prediction and fine-tuning.

**Data Preparation for Fine-Tuning**

During the data preparation phase, the apple and corn datasets, each containing 800 samples, were randomly split into training, validation, and test sets using the same sample method with a fixed random state of 42. The dataset was split in an 80-20 ratio, resulting in 640 training samples and 160 test samples. The training set was further divided into 512 samples for model training and 128 samples for validation. In all subsets, label distribution was maintained equally to ensure balanced representation across all disease categories. This splitting procedure was applied consistently to both apple and corn datasets.

Later in this article, we will examine not only the fine-tuning of the GPT-4o model on the full training set but also the implementation of a progressive fine-tuning approach. This method provides deeper insights into the model's learning capabilities by progressively incorporating new samples into subsequent fine-tuning iterations. To facilitate the progressive fine-tuning process, the training set was further divided into four subsets, each corresponding to a distinct phase of progressive fine-tuning.

*C. Model Selection*

In this study, the primary objective is to identify and classify plant leaf diseases. Numerous models have been employed in prior research to achieve this objective, achieving varying degrees of success. Among the most notable models used for plant disease classification are:

- DenseNet [81]: A densely connected convolutional network that enhances feature propagation and mitigates the vanishing gradient problem.
- ConvNeXt [82]: A convolutional neural network (CNN) inspired by transformer architectures, incorporating design improvements for enhanced efficiency and performance.
- Vision Transformer (ViT) [83]: A transformer-based model that processes images as sequences of patches, utilizing self-attention instead of traditional convolutions.
- ResNet [84, 85]: One of the most well-researched and highly accurate models for this task, ResNet (Residual Network) employs deep residual learning and skip connections, enabling very deep architectures without suffering from vanishing gradients.

For this study, we selected the ResNet architecture due to its strong performance and demonstrated effectiveness in image classification tasks. When fine-tuned for plant disease detection, ResNet has been shown to achieve near-perfect accuracy. The ResNet family includes various architectures with different depths and parameter sizes, such as ResNet-18, ResNet-34, ResNet-50, ResNet-101, and ResNet-152, with parameter counts ranging from approximately 11.7M to 60.2M. Generally, deeper networks capture more complex patterns but also require more computational resources and are more susceptible to overfitting if not adequately regularized. While ResNet-152 provides greater capacity for learning intricate features, its high computational cost and data requirements make it less practical for smaller datasets. Given these trade-offs, we selected ResNet-50, which strikes a balance between model depth and computational efficiency.

**Evaluating Multimodal LLMs**

A key focus of this study is assessing the potential of multimodal LLMs in this task. Specifically, we aim to investigate whether multimodal LLMs such as GPT-4o can accurately classify plant leaf diseases, and if so, at what computational cost and inference time.

LLMs have recently gained widespread adoption in artificial intelligence, surpassing traditional models in various domains. Many tasks previously addressed with lower-parameter models such as RNNs, GRUs, or traditional transformer-based models like BERT can now be tackled more efficiently using LLMs. However, these advancements come at a significant cost: LLMs are pre-trained on billions of parameters, originally optimized for NLP tasks. Their architectural advancements in recent years has led to enhanced capabilities, integrating computer vision, audio processing, reasoning, and more. A defining advantage of LLMs is their ability to perform complex tasks in a zero-shot setting, without explicit task-specific fine-tuning, enabling their broader applicability.

In this study, we benchmark GPT-4o on plant disease classification, evaluating its performance under zero-shot and few-shot learning paradigms. The motivation for this approach stems from OpenAI's claim that GPT-4o can adapt to vision-based tasks with minimal training. According to OpenAI, case studies have demonstrated that "even with just 100 image samples, the GPT-4o multimodal model could significantly improve its performance on vision tasks" [86]. This raises important research questions regarding the feasibility, accuracy, and efficiency of using multimodal LLMs in plant disease detection.

*D. Prompt Engineering*

Prompt engineering is a crucial methodology for optimizing the performance of LLMs across diverse tasks, including few-shot classification [87]. This process involves constructing well-defined input prompts that guide the model toward generating accurate and contextually relevant responses. The effectiveness of a prompt significantly influences the model's ability to interpret task requirements and produce high-quality outputs, making prompt design a critical factor in harnessing the full potential of LLMs.

In this study, prompt engineering played a central role in enabling the model to classify plant leaf images into predefined disease categories without prior training on those specific diseases. Given the adaptability of LLMs, designing effective prompts required an iterative refinement process. This process involved systematic adjustments to improve model comprehension, reduce ambiguity, and enhance classification accuracy. Key considerations included instruction specificity, input data formatting, and overall prompt structure, all of which were carefully optimized to ensure robust performance [88].

To develop an effective prompt for this research, we conducted an iterative trial-and-error process using GPT's chat interface, refining the prompt structure based on empirical results. Recognizing that structured data formats often enhance LLM performance compared to unstructured text, we initially explored XML-based formatting to organize the input. This approach provided a clear representation of image-related information, facilitating better model interpretability. However, despite its benefits, XML formatting increased the prompt's length, leading to a higher token count and, consequently, greater computational cost when making API calls.

To address this challenge, we leveraged Anthropic's console prompt generator, a specialized tool designed to assist in crafting effective prompts by providing structured templates aligned with best practices [89]. Through experimentation, we determined that JSON formatting was a more efficient alternative to XML. JSON not only streamlined the prompt's structure but also reduced token consumption, making API interactions more cost-effective while maintaining model interpretability.

The final prompt was thoroughly refined to balance clarity, efficiency, and computational resource constraints. This structured approach to prompt engineering maximized the model's performance while mitigating the trade-offs associated with API costs and token limitations. **Fig. 4** illustrates the final optimized prompt used in our experiments. By implementing a systematic and data-driven prompt engineering strategy, we effectively harnessed the capabilities of multimodal LLMs for plant disease classification.

```python
type = 'Apple'
image = 'Apple/256/healthy/healthy-1189.JPG'
json = {"categories": ["black-rot", "healthy", "rust", "scab"]}
conversation.append({
        'role': 'user',
        "content": [
            {
                "type": "text",
                "text": 'Analyze the provided image of an ' + type + ' leaf using your computer vision capabilities.'
                        ' Classify the leaf into the most appropriate category based on its condition, choosing from'
                        ' the predefined list: ' + str(json["categories"]) + '. Provide your final classification'
                        ' in the following JSON format without explanations: {"category": "chosen_category_name"}',
            },{
                "type": "image_url",
                "image_url": {
                    "url": image
                }
            }
        ]
})
```

**Fig. 4.** Computer vision JSON-formatted prompt for zero-shot and few-shot classification

### E. Fine-Tuning with Few-Shot Learning

In this study, the fine-tuning process utilized few-shot learning to expose the selected models to previously unseen labeled data. Specifically, 512 plant leaf images per dataset (Apple and Corn) along with their corresponding disease labels were introduced to the models. A validation set consisting of 128 samples was also employed. The same fine-tuning procedure was applied across different image resolutions (100, 150, and 256 pixels). Each fine-tuned model was saved with a descriptive name for inference and subsequently used to generate predictions on a test set comprising 160 samples.

**Fine-Tuning the ResNet-50 Model Using Bayesian Optimization**

Fine-tuning is an iterative process wherein a pre-trained model is adapted to a specific task by adjusting its parameters using labeled data. This process involves configuring hyperparameters such as the number of epochs, batch size, and learning rate. Each of these hyperparameters plays a crucial role in determining model performance. For instance, larger batch sizes can stabilize training and enhance computational parallelism but may increase memory requirements and reduce generalization capabilities [90]. Similarly, higher learning rates can accelerate convergence but may also lead to instability by overshooting the optimal solution [91]. While increasing the number of epochs allows the model to capture complex patterns, it also raises the risk of overfitting [92, 93]. Therefore, selecting optimal hyperparameters is essential for maximizing model performance during both training and inference.

Various methodologies exist for identifying optimal hyperparameters, including Grid Search, Random Search, and Bayesian Optimization. Grid Search is an exhaustive approach that systematically evaluates all possible combinations of specified hyperparameters. While it ensures that the best combination within the predefined space is identified, this method is computationally expensive and time-consuming, particularly when the hyperparameter space is large.

In this study, we employed Bayesian Optimization due to its efficiency in exploring complex search spaces. Bayesian Optimization utilizes a probabilistic model to predict promising hyperparameters based on prior evaluations, thereby reducing the number of function evaluations required [94]. This method iteratively balances exploration (sampling uncertain regions) and exploitation (focusing on known high-performing areas), resulting in more computationally efficient searches compared to Grid Search [95]. This approach is particularly advantageous for fine-tuning models with a large number of parameters, where an exhaustive search would be impractical.

The objective of Bayesian Optimization is to identify the optimal hyperparameters by iteratively refining predictions based on prior evaluations. Specifically, we implemented the Tree-structured Parzen Estimator (TPE) algorithm, which follows these general steps:

- Modeling the Objective Function: The objective function $f(x)$, represents model performance as a function of the hyperparameters $x$. The goal is to identify the optimal set of hyperparameters $x^*$ that minimizes or maximizes $f(x)$.
- Probabilistic Model: TPE approximates the distribution of objective function values using a probabilistic model, which can be represented as:

$$p(y|x) \sim N(\mu(x), \sigma(x)) \quad (1)$$

- Surrogate Function: The TPE algorithm constructs a surrogate model to select the next set of hyperparameters based on the posterior probability distribution. The expected improvement (EI) criterion is often used to guide this selection:

$$EI(x) = E[\max(f(x^*) - f(x), 0)] \quad (2)$$

- TPE estimates the probability of a candidate hyperparameter improving performance by modeling the density ratio between high-performing and low-performing observations:

$$TPE(x) = \frac{p(x|y < y_{best})}{p(x|y \geq y_{best})} \quad (3)$$

To implement Bayesian Optimization, we utilized the Optuna library with the TPE sampler. The optimization process was conducted on image datasets of plant leaves, including both training and validation samples with corresponding disease labels, as mentioned before. Optuna's trial-based framework captured validation accuracy at each epoch, enabling dynamic updates to its probabilistic model and more accurate hyperparameter recommendations. Additionally, the library's early stopping and trial pruning features were leveraged to terminate underperforming trials, enhancing computational efficiency. All procedures were executed on Google Colab with an NVIDIA A100-SXM4-40GB GPU.

Following 30 optimization trials, the best-performing hyperparameters were identified. The fine-tuned ResNet-50 model was saved in .pth format for future inference. This methodology was independently applied to both Apple and Corn leaf datasets across multiple image resolutions. The resulting models were optimized for accurate disease classification, specific to each crop and resolution.

### Fine-Tuning the GPT-4o Multimodal LLM

GPT-4o, OpenAI's flagship multimodal LLM, operates as a closed-source model accessible via an API. Given the associated costs of both inference and fine-tuning, extensive hyperparameter optimization using methods such as Grid Search or Bayesian Optimization can incur substantial computational expenses.

In this study, we hypothesized that the optimal hyperparameters identified during the ResNet-50 fine-tuning process could yield comparable performance when applied to the GPT-4o model. While this assumption carries the risk of suboptimal outcomes due to architectural differences, adopting these pre-optimized hyperparameters offers a pragmatic approach to control computational costs.

Fine-tuning the GPT-4o model required preparing training data in JSONL format, consisting of prompt-completion pairs. Separate training and validation files were generated for each crop (Apple and Corn) and each image resolution (100px, 150px, and 256px), resulting in six training files and six validation files. These files were derived directly from the original CSV datasets. **Fig. 5** provides an example of a JSONL training file, illustrating the prompt-completion structure.

Using these data files, we initiated six fine-tuning jobs via the GPT-4o API, applying the hyperparameters identified during the ResNet-50 optimization. Upon completion, the fine-tuned models were used for prediction tasks, yielding crop- and resolution-specific outputs. This streamlined approach allowed us to assess the transferability of hyperparameters across different model architectures while maintaining computational efficiency.

### Progressive Fine-Tuning of GPT-4o

To further investigate the applicability of hyperparameters derived from the Bayesian optimization of the ResNet-50 model, we adopted a progressive fine-tuning methodology for the GPT-4o model.

In this methodology, we allowed the OpenAI platform to select the hyperparameters for epochs, learning rate, and batch size by leaving these inputs at their default settings. The platform chose a conservative yet cost-efficient configuration, with a batch size of 1 and 3 training epochs.

Progressive fine-tuning was conducted in four phases. In each phase, the previously fine-tuned model was further refined using an additional 128 samples. This phased approach provided insights into the suitability of the default hyperparameters, as well as the model's capacity to adapt to incremental training data.

```
{"messages": [
{"role": "user", "content":
    "Analyze the provided image of an apple leaf using your computer vision capabilities. Classify the"
    "leaf into the most appropriate category based on its condition, choosing from the predefined list:"
    "{\n  \"categories\": [\n    \"black-rot\",\n    \"healthy\",\n    \"rust\",\n    \"scab\"\n  ]\n}"
    "Provide your final classification in the following JSON format without explanations:"
    "{\"category\": \"chosen_category_name\"}"},
{"role": "user", "content": [
    {"type": "image_url", "image_url": {"url": "256/black-rot/black-rot-256.JPG"}}
]},
{"role": "assistant", "content": "{\"category\": \"black-rot\"}"}]}
```

**Fig. 5.** Prompt and completion pairs – JSONL files for fine-tuning the GPT-4o multimodal LLM on image data.

As a result of this process, we produced 12 fine-tuned GPT-4o models for each crop and resolution. Each model was subsequently used to generate predictions for the test sets, allowing us to assess performance across different stages of the progressive fine-tuning process.

*F. Model Evaluation Strategy*

During the evaluation phase, the fine-tuned GPT-4o and ResNet-50 models were utilized to generate predictions on the test datasets. Each model was provided with test leaf images and tasked with predicting the corresponding plant disease based on its training. The predictions from each model were recorded in separate columns for subsequent analysis, while the predictions from the ResNet-50 model were also stored in separate files for organizational clarity.

To evaluate the performance of the models in the plant disease detection and classification task, standard evaluation metrics were employed, including accuracy, precision, recall, F1 score, and heatmaps. These metrics were applied consistently across all predictions to facilitate direct comparison between the models.

$$Accuracy = \frac{TP + TN}{TP + TN + FP + FN} \quad (4)$$

$$Recall = \frac{TP}{TP + FN} \quad (5)$$

$$Precision = \frac{TP}{TP + FP} \quad (6)$$

$$F1 = 2 \, x \, \frac{Precision \, x \, Recall}{Precision + Recall} \quad (7)$$

Additional cross-resolution prediction experiments were conducted to further investigate the GPT-4o model's computer vision capabilities. In these experiments, models fine-tuned on lower-resolution images were used to predict diseases on higher-resolution test sets, and vice versa. This approach enabled the assessment of the models' adaptability and robustness across varying image resolutions, providing a more comprehensive understanding of their performance in real-world applications.

Finally, the same evaluation methodology was applied to the best-performing model at each resolution to predict diseases on a different crop. This additional experiment aimed at assessing the generalization ability of the fine-tuned models across different crop types, offering insights into their broader applicability beyond the training dataset.

IV. RESULTS

This section presents the outcomes of our experiments. The primary objective was to evaluate the effectiveness of the GPT-4o multimodal LLM in detecting and classifying plant leaf diseases, both before and after fine-tuning using few-shot learning. For direct comparison, the ResNet-50 model was also fine-tuned using the same methodology. As outlined in the methodology section, a structured evaluation process was implemented. This structure will be maintained throughout the presentation of the findings for clarity.

We begin by comparing the results of fine-tuning on the complete training sets, followed by an analysis of progressive fine-tuning, cross-resolution evaluations, and cross-plant evaluations. The results, in terms of various performance measures described in the methodology section, are presented in the following subsections, focusing on training efficiency, computational cost, and classification accuracy.

*A. GPT-4o and ResNet-50 Fine-Tuning*

As detailed in the methodology section, both the GPT-4o multimodal LLM and the ResNet-50 model underwent fine-tuning with few-shot learning to improve their ability to accurately detect and classify plant leaf diseases. The fine-tuning process utilized a training set of 512 labeled images and a validation set of 128 images. Hyperparameters were optimized using Bayesian optimization on the ResNet-50 model, and the same optimized parameters were applied to both models.

Fine-tuning was conducted on two plant species (apple and corn) and across three image resolutions (100, 150, and 256 pixels). Table I provides a detailed summary of the metrics obtained during the fine-tuning process, including training and validation loss, training time, and associated computational costs. These results offer a comprehensive evaluation of the models' performance across different image resolutions, highlighting their scalability and efficiency in plant leaf disease detection and classification.

TABLE I
FINE-TUNING METRICS AND COMPUTATIONAL COSTS FOR GPT-4O AND RENNET-50
FINE-TUNING ACROSS DIFFERENT IMAGE RESOLUTIONS

| Plant | Subset Identifier | Epochs | Batch Size | Training Loss | Full Validation Loss | Training Duration (e.g., Seconds) | Training Cost (e.g., USD) | Base Model | Output Model |
|---|---|---|---|---|---|---|---|---|---|
| Apple | GPT-Resolution-256 | 10 | 16 | 0.0000 | 0.0088 | 1,778.0 | 47.5300 | gpt-4o-2024-08-06 | BCjDLKKo |
| | ResNet-Resolution-256 | 10 | 16 | 0.3497 | 0.1286 | 280.2 | 0.0708 | resnet-50 | apple-best_resnet50_model-256.pth |
| | GPT-Resolution-150 | 10 | 16 | 0.0000 | 0.0015 | 1,732.0 | 47.4600 | gpt-4o-2024-08-06 | BCsfA4QN |
| | ResNet-Resolution-150 | 10 | 16 | 0.5987 | 0.3846 | 253.2 | 0.0640 | resnet-50 | apple-best_resnet50_model-150.pth |
| | GPT-Resolution-100 | 10 | 16 | 0.0000 | 0.0155 | 2,172.0 | 47.7100 | gpt-4o-2024-08-06 | BCtaE9Ou |
| | ResNet-Resolution-100 | 10 | 16 | 0.6094 | 0.3676 | 241.2 | 0.0609 | resnet-50 | apple-best_resnet50_model-100.pth |
| Corn | GPT-Resolution-256 | 10 | 16 | 0.0000 | 0.0286 | 1,679.0 | 47.0200 | gpt-4o-2024-08-06 | BCuf5MWW |
| | ResNet-Resolution-256 | 10 | 16 | 0.4276 | 0.2814 | 261.0 | 0.0660 | resnet-50 | corn-best_resnet50_model-256.pth |
| | GPT-Resolution-150 | 10 | 16 | 0.0000 | 0.0267 | 1,909.0 | 45.3500 | gpt-4o-2024-08-06 | BD6xrnn3 |
| | ResNet-Resolution-150 | 10 | 16 | 0.6150 | 0.5232 | 247.2 | 0.0625 | resnet-50 | corn-best_resnet50_model-150.pth |
| | GPT-Resolution-100 | 10 | 16 | 0.0000 | 0.0358 | 2,109.0 | 48.3900 | gpt-4o-2024-08-06 | BD7xmXAh |
| | ResNet-Resolution-100 | 10 | 16 | 0.5827 | 0.3615 | 232.2 | 0.0587 | resnet-50 | corn-best_resnet50_model-100.pth |

Based on the results in Table I, the GPT-4o models consistently achieved lower training and validation losses across all subsets compared to the ResNet-50 models, indicating superior accuracy and better generalization capabilities. Despite their higher computational costs —ranging from $45.35 to $48.39 when fine-tuned on the corn image samples and $47.46 to $47.71 when fine-tuned on the apple samples—and longer training durations (approximately 1,679 to 2,172 seconds), the GPT-4o models demonstrated near-zero training loss and minimal validation loss. In contrast, the ResNet-50 models exhibited higher training and validation losses across all resolutions, with these losses increasing as the resolution decreased. However, the ResNet-50 models required significantly less time (approximately 232.2 to 280.2 seconds) and incurred substantially lower training costs —ranging from $0.0587 to $0.0660 when fine-tuned on corn image samples and $0.0609 to $0.0708 when fine-tuned on apple image samples. These results suggest that while the GPT-4o models offer improved accuracy and generalization, they do so at the expense of increased computational resources. Notably, the performance gap between the two model architectures is more pronounced at lower resolutions, where the ResNet-50 models exhibit greater difficulty in maintaining low validation loss. It is important to acknowledge that these differences in performance are also influenced by the distinct architectural frameworks of GPT-4o and ResNet-50, meaning that variations in training and validation loss cannot be fully attributed to a single factor, and discrepancies may arise during the prediction phase.

During the GPT-4o fine-tuning process, unexpected errors were encountered. Prior to fine-tuning, OpenAI performs an automated review by downloading and scanning the training and validation images to ensure compliance with their terms of use. Errors occurred during the fine-tuning of the GPT-4o model on the apple training set with resolutions of 256 and 150, where the system flagged 7 samples as containing prohibited content. Specifically, OpenAI's detection mechanism identified images as containing faces, people, or CAPTCHA-like patterns. This issue was even more pronounced in the corn training set, where the system flagged and excluded 17, 35, and 3 samples for the 256, 150, and 100 resolutions, respectively, due to the detection of human-like content. As a result, these samples were automatically removed from the training sets to allow the fine-tuning process to continue.

These unexpected detections raise concerns regarding the accuracy and sensitivity of OpenAI's content detection mechanism, particularly how it identified human-related content in datasets consisting solely of plant imagery. It remains unclear whether these detections were triggered by reflections or patterns resembling human faces, or if the detection algorithm requires further calibration. The frequency of these erroneous detections, especially in the corn training set, suggests a potential misalignment between the content detection system and the nature of agricultural images. This issue warrants further investigation, which will be closely monitored in the progressive fine-tuning section.

### B. Prediction Performance on Test Sets

During the prediction phase, each fine-tuned model was evaluated on its corresponding test set, aiming to detect diseases in plant leaf image samples. The results of these evaluations are summarized in Table II.

TABLE II
FEW-SHOT IMAGE CLASSIFICATION PERFORMANCE OF GPT-4O AND RESNET-50 FINE-TUNED MODELS

| PLANT | PREDICTION COLUMN | FINE-TUNED MODEL | ACCURACY | PRECISION | RECALL | F1 | PREDICTION DURATION (SECONDS) | PREDICTION COST (USD) |
|---|---|---|---|---|---|---|---|---|
| Apple | GPT-Resolution-256 | BCjDLKKo | 0.9812 | 0.9820 | 0.9812 | 0.9811 | 579.85 | 0.2300 |
| | ResNet-Resolution-256 | apple-best_resnet50_model-256.pth | 0.9688 | 0.9710 | 0.9688 | 0.9690 | 28.22 | 0.0079 |
| | GPT-Resolution-150 | BCsfA4QN | 0.9563 | 0.9568 | 0.9562 | 0.9562 | 567.07 | 0.2300 |
| | ResNet-Resolution-150 | apple-best_resnet50_model-150.pth | 0.9250 | 0.9308 | 0.9250 | 0.9251 | 23.19 | 0.0065 |
| | GPT-Resolution-100 | BCtaE9Ou | 0.9563 | 0.9596 | 0.9562 | 0.9564 | 572.20 | 0.2300 |
| | ResNet-Resolution-100 | apple-best_resnet50_model-100.pth | 0.8500 | 0.8576 | 0.8500 | 0.8488 | 20.38 | 0.0057 |
| Corn | GPT-Resolution-256-Corn | BCuf5MWW | 0.9187 | 0.9227 | 0.9188 | 0.9189 | 586.72 | 0.2300 |
| | ResNet-Resolution-256 | corn-best_resnet50_model-256.pth | 0.8750 | 0.8778 | 0.8750 | 0.8752 | 23.09 | 0.0065 |
| | GPT-Resolution-150-Corn | BD6xrnn3 | 0.8812 | 0.8861 | 0.8812 | 0.8782 | 595.87 | 0.2300 |
| | ResNet-Resolution-150-Corn | corn-best_resnet50_model-150.pth | 0.8000 | 0.8101 | 0.8000 | 0.7959 | 18.19 | 0.0051 |
| | GPT-Resolution-100-Corn | BD7xmXAh | 0.9313 | 0.9320 | 0.9312 | 0.9305 | 574.37 | 0.2300 |
| | ResNet-Resolution-100-Corn | corn-best_resnet50_model-100.pth | 0.8063 | 0.8042 | 0.8062 | 0.8038 | 25.24 | 0.0071 |

As observed in Table II for both the Apple and Corn datasets, GPT-4o consistently outperforms ResNet-50 across all image resolutions in terms of classification accuracy, precision, recall, and F1-score. Specifically, GPT-4o achieves the highest accuracy of 0.9812 on the Apple dataset at a resolution of 256, compared to ResNet-50's accuracy of 0.9688 at the same resolution. However, as image resolution decreases, both models experience a decline in performance, with ResNet-50 showing a more pronounced drop, particularly at a resolution of 100, where its accuracy falls to 0.8500 for Apple and 0.8063 for Corn. GPT-4o maintains relatively higher precision and recall across all resolutions, reinforcing its robustness in classification tasks.

Nevertheless, this enhanced performance comes with significant trade-offs in terms of computational cost and prediction time, similar to what was observed during model training. GPT-4o requires substantially more time for inference, with total durations ranging from 574.37 to 596.72 seconds for 160 plant leaf images—equivalent to approximately 3.59 to 3.73 seconds per image. In contrast, ResNet-50 completes predictions significantly faster, ranging from 20.38 to 28.22 seconds for Apple and 18.19 to 25.24 seconds for Corn, which translates to approximately 0.13 to 0.18 seconds per image. Furthermore, GPT-4o incurs a fixed cost of $0.23 per prediction phase (for 160 plant leaf images), equating to roughly $0.00144 per image. ResNet-50, by comparison, operates at a much lower cost—$0.0057 to $0.0079 for Apple and $0.0051 to $0.0071 for Corn in total—corresponding to approximately $0.000036 to $0.000049 per image. These results highlight a critical trade-off: while GPT-4o delivers better classification performance, it demands higher computational resources and incurs substantially greater costs. Conversely, ResNet-50, despite its lower accuracy, provides significant advantages in terms of speed and cost-efficiency, making it a more practical choice for real-time applications with resource constraints.

An interesting observation is that both models achieve lower accuracy on the Corn dataset compared to the Apple dataset. Specifically, GPT-4o attains an accuracy of 91.87% for Corn, compared to 98.12% for Apple. The similar decline in performance across both models suggests that disease classification in corn leaves may be more challenging, potentially due to the complexity of leaf structures or the visual similarity of disease symptoms, making differentiation more difficult. This finding implies that the model encounters greater difficulty distinguishing between disease patterns in corn leaves, whereas the disease features in apple leaves may be more distinct.

Another notable observation is the unexpected decline in GPT-4o's performance at a resolution of 150 compared to 100, where it exhibits higher accuracy. This drop can likely be attributed to the exclusion of certain training images specifically, 35 images were removed due to OpenAI's automated system flagging them as containing human faces. The removal of these images may have reduced dataset diversity, thereby negatively impacting the model's generalization capability. Additionally, the lower 91.87% accuracy at a resolution of 256 for Corn, compared to 98.12% for Apple, may also be linked to the exclusion of 17 image samples from the training set, which were similarly flagged as containing human faces. The exclusion of these training images may have eliminated valuable visual features necessary for distinguishing between disease patterns in corn leaves. These exclusions suggest that image content flagged by automatic systems, such as human faces, can significantly impact the model's learning process and, consequently, its classification accuracy. If this issue cannot be addressed by OpenAI, future research could begin progressive fine-tuning with a larger dataset, consistently removing the same number of flagged images from all datasets to maintain balance and comparability.

### C. GPT-4o Progressive Fine-Tuning Process

The observations in the previous section raised two key questions: (1) whether the hyperparameters optimized for ResNet-50 fine-tuning via Bayesian optimization are also effective for fine-tuning GPT-4o, and (2) whether the flagged images represent actual dataset errors, such as containing human faces or captchas, or if they were incorrectly identified due to a limitation in OpenAI's filtering mechanism. These uncertainties prompted further investigation through a progressive fine-tuning approach for GPT-4o.

In this phase, the multimodal GPT-4o model undergoes progressive fine-tuning using batches of 128 samples, as described in the methodology section. The fine-tuned model from the initial training phase is further refined through additional fine-tuning cycles. The default hyperparameters recommended by OpenAI's platform, three training epochs with a batch size of one were adopted to assess their suitability for optimizing LLM performance in this domain.

The training metrics and evaluation results obtained from this experiment are presented in Tables III and VII.A1.

A comparative analysis of Table VII.A1, which presents the prediction results after progressive fine-tuning using OpenAI's default hyperparameters, with Table II, which reports results from fine-tuning the model on the entire training set in a single pass using hyperparameters optimized via Bayesian techniques for ResNet-50, reveals several insights. The highest accuracy in Table II was achieved at a resolution of 256 pixels (98.12%), whereas in Table VII.A1, the highest accuracy was observed during phase 3 of fine-tuning at the same resolution (95.63%). Similar trends were noted for images at 150px and 100px resolutions. These results confirm that the hyperparameters optimized for ResNet-50 through Bayesian methods also yield effective results for fine-tuning GPT-4o. This is a significant observation, as applying hyperparameter optimization directly to GPT-4o would incur substantial computational costs for each iteration.

TABLE III
PROGRESSIVE FINE-TUNING METRICS AND COMPUTATIONAL COSTS FOR GPT-4O

| Plant | Subset Identifier | Trained Tokens | Training Loss | Full Validation Loss | Training Duration (e.g., Seconds) | Training Cost (e.g., USD) | Base Model | Output Model | Errors |
|---|---|---|---|---|---|---|---|---|---|
| Apple | Phase-1-Resolution-256 | 143,136 | 0.0000 | 0.1363 | 1,722 | 3.58 | gpt-4o-2024-08-06 | B9ojfjfp | 0 |
| | Phase-2-Resolution-256 | 95,055 | 0.0001 | 0.0648 | 1,501 | 2.38 | B9ojfjfp | B9rgDgVU | 43 |
| | Phase-3-Resolution-256 | 142,014 | 0.0000 | 0.0496 | 2,213 | 3.55 | B9rgDgVU | B9wnegyN | 1 |
| | Phase-4-Resolution-256 | 143,136 | 0.0001 | 0.0617 | 1,773 | 3.58 | B9wnegyN | B9zes8zI | 0 |
| | Phase-1-Resolution-150 | 143,136 | 0.0000 | 0.1572 | 1,772 | 3.58 | gpt-4o-2024-08-06 | BA0XO3aS | 0 |

| | Phase-Resolution | | | | | | | | |
|---|---|---|---|---|---|---|---|---|---|
| | Phase-2-Resolution-150 | 142,020 | 0.0001 | 0.1429 | 2,026 | 3.57 | BA0XO3aS | BA1jLY4h | 1 |
| | Phase-3-Resolution-150 | 142,020 | 0.0002 | 0.0883 | 2,065 | 3.57 | BA1jLY4h | BA6q0cyK | 1 |
| | Phase-4-Resolution-150 | 140,904 | 0.0000 | 0.0983 | 2,110 | 3.52 | BA6q0cyK | BA7oD2kT | 2 |
| | Phase-1-Resolution-100 | 143,136 | 0.0000 | 0.2313 | 1,771 | 3.58 | gpt-4o-2024-08-06 | BACKwp0N | 0 |
| | Phase-2-Resolution-100 | 143,136 | 0.0008 | 0.3491 | 2,057 | 3.57 | BACKwp0N | BAEfIBXZ | 0 |
| | Phase-3-Resolution-100 | 143,136 | 0.0000 | 0.1286 | 1,772 | 3.58 | BAEfIBXZ | BAJW3bri | 0 |
| | Phase-4-Resolution-100 | 138,654 | 0.0000 | 0.1367 | 1,773 | 3.47 | BAJW3bri | BAMVk7Cm | 4 |
| Corn | Phase-1-Resolution-256 | 130,029 | 0.0000 | 0.0837 | 2,162 | 3.26 | gpt-4o-2024-08-06 | BAWXbcjw | 14 |
| | Phase-2-Resolution-256 | 142,599 | 0.0000 | 0.0739 | 1,720 | 3.56 | BAWXbcjw | BAXxy46P | 3 |
| | Phase-3-Resolution-256 | 136,878 | 0.0001 | 0.0672 | 1,672 | 3.42 | BAXxy46P | BAZKgwpN | 8 |
| | Phase-4-Resolution-256 | 142,569 | 0.0004 | 0.0489 | 1,907 | 3.57 | BAZKgwpN | BAaQ3wfm | 3 |
| | Phase-1-Resolution-150 | 139,200 | 0.0002 | 0.1351 | 1,727 | 3.48 | gpt-4o-2024-08-06 | BAc2StJe | 6 |
| | Phase-2-Resolution-150 | 135,738 | 0.0002 | 0.0565 | 1,622 | 3.39 | BAc2StJe | BB8YhNDx | 9 |
| | Phase-3-Resolution-150 | 141,444 | 0.0002 | 0.0368 | 3,118 | 3.53 | BB8YhNDx | BB9ux708 | 4 |
| | Phase-4-Resolution-150 | 134,601 | 0.0002 | 0.0270 | 1,617 | 3.37 | BB9ux708 | BBAo9aAA | 10 |
| | Phase-1-Resolution-100 | 146,016 | 0.0000 | 0.0672 | 2,160 | 3.65 | gpt-4o-2024-08-06 | BBBljn0v | 0 |
| | Phase-2-Resolution-100 | 144,882 | 0.0000 | 0.0625 | 2,259 | 3.62 | BBBljn0v | BBD8rd0v | 1 |
| | Phase-3-Resolution-100 | 142,590 | 0.0005 | 0.0292 | 1,722 | 3.57 | BBD8rd0v | BBDySt2J | 3 |
| | Phase-4-Resolution-100 | 146,016 | 0.0000 | 0.0307 | 2,264 | 3.65 | BBDySt2J | BBF7QPJj | 0 |

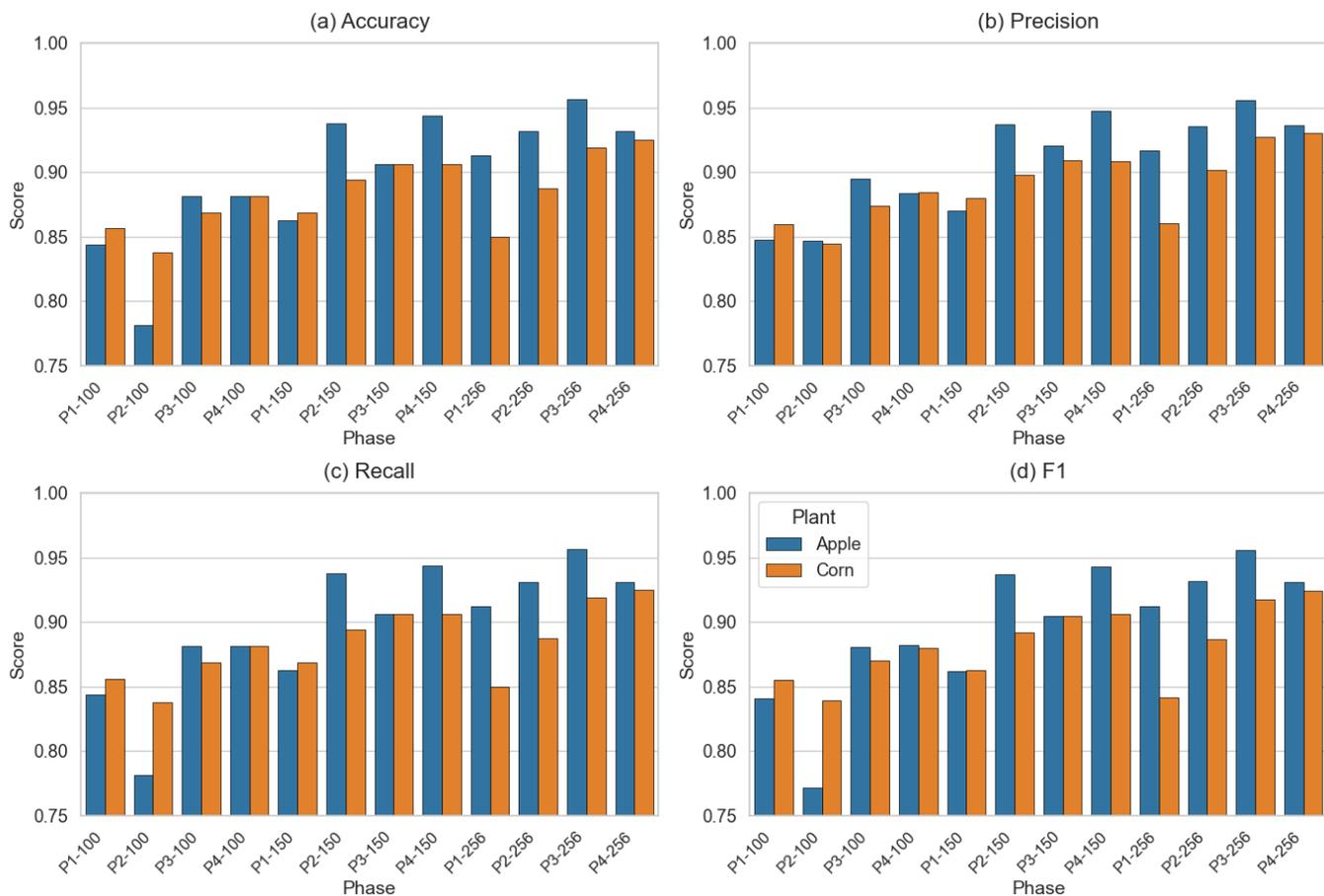

**Fig. 6.** Few-shot classification performance of progressively fine-tuned GPT-4o models. Comparison across training phases, image resolutions, and plant sample types: (a) Accuracy, (b) Precision, (c) Recall, (d) F1 Score.

Furthermore, Table A1 highlights a clear overfitting trend in the GPT-4o model beyond phase 3, across all resolutions. Specifically, for 256px resolution images, accuracy improved up to phase 3 (95.63%) before declining to 93.13% in subsequent phases. This overfitting may be attributed to the default hyperparameters recommended by OpenAI. In contrast, use of the hyperparameters derived from Bayesian optimization of ResNet-50 minimized overfitting, suggesting that practitioners aiming to fine-tune GPT models on image data can first optimize hyperparameters on a ResNet model and then apply them to GPT-4o for improved performance.

An additional observation from Table III pertains to the discrepancies in trained token counts, training costs, and processing time. These variations stem from OpenAI's review mechanism, which excluded a significant number of images from the training dataset as discussed before, marking them as containing people, faces, or captchas. The number of images flagged during each fine-tuning job is recorded in the final column of Table III.

A particularly noteworthy inconsistency arises in OpenAI's image validation mechanism. In the initial fine-tuning experiment (Section IV.A), the review system flagged seven images in the apple training set and 55 in the corn dataset. However, in the current progressive fine-tuning process, the same datasets resulted in 52 flagged images for apples and 61 for corn. This discrepancy suggests that OpenAI's image validation algorithm lacks repeatability, potentially obstructing the model from being trained on the full dataset and thus limiting fine-tuning effectiveness.

### D. Zero-Shot Classification

As previously mentioned, the GPT-4o model is a pre-trained language model capable of producing accurate results both in fine-tuned and zero-shot settings. To evaluate its zero-shot performance, an independent test was conducted using the GPT-4o base model and its mini variant. In this test, the models were prompted to classify test set images to their corresponding plant diseases without prior exposure to the specific task.

TABLE IV
ZERO-SHOT IMAGE CLASSIFICATION PERFORMANCE
OF GPT-4O AND GPT-4O MINI BASE MODELS ON THE SAME TEST SETS

| PLANT | PREDICTION COLUMN | BASE MODELS | ACCURACY | PRECISION | RECALL | F1 | PREDICTION DURATION (E.G., SECONDS) | PREDICTION COST (E.G., USD) |
|---|---|---|---|---|---|---|---|---|
| Apple | GPT-4o-Resolution-256 | gpt-4o | 0.5687 | 0.5616 | 0.5688 | 0.5031 | 386.94 | 0.15 |
| | GPT-4o-mini-Resolution-256 | gpt-4o-mini | 0.5188 | 0.5007 | 0.5188 | 0.4595 | 399.01 | 0.11 |
| | GPT-4o-Resolution-150 | gpt-4o | 0.5188 | 0.4944 | 0.5188 | 0.4508 | 381.09 | 0.14 |
| | GPT-4o-mini-Resolution-150 | gpt-4o-mini | 0.4813 | 0.4697 | 0.4813 | 0.4032 | 387.67 | 0.11 |
| | GPT-4o-Resolution-100 | gpt-4o | 0.4688 | 0.4385 | 0.4688 | 0.3874 | 382.81 | 0.14 |
| | GPT-4o-mini-Resolution-100 | gpt-4o-mini | 0.4313 | 0.3782 | 0.4313 | 0.3217 | 366.32 | 0.10 |
| Corn | GPT-4o-Resolution-256 | gpt-4o | 0.6875 | 0.6646 | 0.6875 | 0.6457 | 516.00 | 0.15 |
| | GPT-4o-mini-Resolution-256 | gpt-4o-mini | 0.6375 | 0.5828 | 0.6375 | 0.5955 | 448.19 | 0.11 |
| | GPT-4o-Resolution-150 | gpt-4o | 0.6937 | 0.7062 | 0.6937 | 0.6461 | 467.43 | 0.07 |
| | GPT-4o-mini-Resolution-150 | gpt-4o-mini | 0.6687 | 0.6141 | 0.6687 | 0.6054 | 415.24 | 0.01 |
| | GPT-4o-Resolution-100 | gpt-4o | 0.6375 | 0.6359 | 0.6375 | 0.5933 | 390.21 | 0.08 |
| | GPT-4o-mini-Resolution-100 | gpt-4o-mini | 0.6188 | 0.6147 | 0.6188 | 0.5852 | 311.09 | 0.01 |

The zero-shot classification results, presented in Table IV, were notably lower compared to those obtained after fine-tuning. Specifically, the GPT-4o base model and its mini variant achieved maximum accuracies of 56.87% and 51.88%, respectively, on the Apple test set, and 69.37% and 66.87%, respectively, on the Corn test set. The models demonstrated higher accuracy in identifying diseases on corn leaves compared to apple leaves in the zero-shot setting, which was reversed after fine-tuning, where the models achieved 98.12% accuracy on apple leaves at a resolution of 256 and 91.87% on corn leaves. Fine-tuning resulted in a substantial performance improvement of 41.25% for apple leaf disease classification and 22.5% for corn leaf disease classification.

These findings demonstrate that a general-purpose LLM can be effectively adapted to a new task with minimal training through few-shot learning, achieving significantly improved performance.

### E. Cross-Resolution Classification and Evaluation

An additional aspect explored in this study is the ability of fine-tuned GPT models to effectively classify plant diseases across images of varying resolutions. To assess this capability, a cross-resolution evaluation was conducted, with the results presented in Table V.

TABLE V
CROSS-RESOLUTION IMAGE CLASSIFICATION PERFORMANCE

| PREDICTION COLUMN | BEST MODEL | ACCURACY | PRECISION | RECALL | F1 | PREDICTION DURATION (E.G., SECONDS) | PREDICTION COST (E.G., USD) |
|---|---|---|---|---|---|---|---|
| Corn-High-to-Low-Res-Trained-256-Prediction-100 | BAaQ3wfm | 0.8500 | 0.8545 | 0.8500 | 0.8470 | 700.38 | 0.23 |
| Corn-Low-to-High-Res-Trained-100-Prediction-256 | BBF7QPJj | 0.9250 | 0.9288 | 0.9250 | 0.9246 | 723.74 | 0.23 |
| Apple-High-to-Low-Res-Trained-256-Prediction-100 | B9wnegyN | 0.7250 | 0.7694 | 0.7250 | 0.7170 | 582.33 | 0.23 |
| Apple-Low-to-High-Res-Trained-100-Prediction-256 | BAJW3bri | 0.8625 | 0.9113 | 0.8625 | 0.8513 | 616.05 | 0.23 |

It is found that models fine-tuned on lower-resolution images and tested on higher-resolution images achieved higher accuracy compared to models trained on high-resolution images and tested on lower-resolution images. For instance, the Corn-Low-to-High-Res-Trained-100-Prediction-256 model achieved a high accuracy of 92.5%, notably outperforming its high-to-low counterpart at 85.0%. Similarly, for apples, the low-to-high model reached 86.25% accuracy compared to 72.5% for the high-to-low model.

This suggests that models trained on lower-resolution data may develop a more refined ability to capture essential features, whereas models trained on high-resolution images struggle to generalize effectively when presented with lower-resolution counterparts.

Interestingly, despite these performance differences, all models maintained comparable prediction durations (~580–720 seconds) and uniform cost estimates (~$0.23 per run), indicating that improvements in accuracy from resolution-

aware training strategies can be achieved without additional computational or financial overhead. This highlights the practicality of using low-to-high resolution training as an efficient and cost-effective approach for real-world deployment.

*F. Evaluation of Cross-Plant Classification and Generalization*

This study concludes with an additional evaluation assessing the generalization capabilities of LLMs in plant disease detection. Specifically, we examine whether models fine-tuned on disease detection for apple images can generalize to identifying diseases in other plant species, such as corn. The results of this cross-plant classification evaluation are presented in Table VI.

TABLE VI
CROSS-PLANT IMAGE CLASSIFICATION PERFORMANCE
BEST MODEL AT EACH RESOLUTION MAKING PREDICTIONS ON A DIFFERENT PLANT TO EVALUATE GENERALIZATION

| PREDICTION COLUMN | BEST MODEL | ACCURACY | PRECISION | RECALL | F1 | PREDICTION DURATION (E.G., SECONDS) | PREDICTION COST (E.G., USD) |
|---|---|---|---|---|---|---|---|
| Best-Corn-Trained-Model-Predictions-on-Apples-256 | BAaQ3wfm | 0.6250 | 0.6781 | 0.6250 | 0.5712 | 618.92 | 0.23 |
| Best-Corn-Trained-Model-Predictions-on-Apples-150 | BB9ux708 | 0.5687 | 0.5684 | 0.5687 | 0.5102 | 605.82 | 0.23 |
| Best-Corn-Trained-Model-Predictions-on-Apples-100 | BBF7QPJj | 0.6062 | 0.5480 | 0.6062 | 0.5433 | 599.37 | 0.23 |
| Best-Apple-Trained-Model-Predictions-on-Corns-256 | B9wnegyN | 0.6687 | 0.6431 | 0.6687 | 0.6415 | 643.49 | 0.23 |
| Best-Apple-Trained-Model-Predictions-on-Corns-150 | BA7oD2kT | 0.4938 | 0.6016 | 0.4938 | 0.4163 | 628.43 | 0.23 |
| Best-Apple-Trained-Model-Predictions-on-Corns-100 | BAJW3bri | 0.5500 | 0.5120 | 0.5500 | 0.4994 | 609.32 | 0.23 |

The findings from this evaluation clearly indicate that fine-tuned models struggle to generalize their classification capabilities across different plant species. While models trained on apple images exhibit some level of disease detection when applied to corn, and vice versa, the overall performance remains suboptimal.

It was found that the highest-performing model in cross-plant classification, the Best-Apple-Trained-Model-Predictions-on-Corns-256, achieved an accuracy of 66.87%, while other models demonstrated substantially lower performance. The decrease in accuracy and F1 scores suggests that the learned features from one plant species are not directly transferable to another, likely due to morphological and textural differences between apple and corn diseases.

Previous research supports this conclusion. For instance, some studies employing the Inception architecture have demonstrated success in detecting the presence of leaf infections across crop types without requiring prior knowledge of the specific plant or disease. However, these studies typically address a simplified binary classification task (Healthy vs. Unhealthy), which is significantly less complex than the four-class disease-specific classification explored in our work [96].

A potentially more robust approach would involve collecting representative samples for each plant species and each associated disease, then training a general model capable of classifying diseases across multiple crops. While such a strategy is beyond the scope of this study, it could offer a path toward broader generalization. However, implementing this approach, particularly when leveraging multimodal LLMs or other deep architectures would likely be computationally intensive and require substantial resources. Given these constraints, achieving results comparable to those produced by specialized, species-specific disease detection models, such as those presented in this study, may not be feasible without significant investment in data curation and model optimization.

V. DISCUSSION AND FUTURE DIRECTIONS

In the preceding sections, the computer vision classification capabilities of the GPT-4o model were systematically evaluated for plant disease detection using leaf images. For direct comparison, the ResNet-50 model was selected as a baseline. Bayesian optimization was applied to the ResNet-50 model to determine the optimal hyperparameters for this task. The same set of hyperparameters was then employed for fine-tuning the GPT-4o model, yielding promising results, with accuracy reaching up to 98.12%.

To assess the hypothesis that hyperparameters derived from the Bayesian optimization of the ResNet-50 model could enhance the fine-tuning process of GPT-4o, an additional progressive fine-tuning experiment was conducted. In this phase, the model was incrementally fine-tuned using batches of new samples at each iteration, but with default hyperparameters. The results indicate that the optimized hyperparameters provided an accuracy improvement of up to 2.49% over the default settings suggested by OpenAI.

Given the strong zero-shot capabilities of GPT-based models, a separate evaluation was performed using the base GPT-4o and its smaller variant, GPT-4o-mini, in a zero-shot setting on the same test sets. The findings reveal that while zero-shot classification is considerably less effective than fine-tuned approaches, the models exhibit high adaptability when provided with a small number of labeled examples. Specifically, few-shot fine-tuning improved classification accuracy by up to 41.25%.

Additionally, cross-resolution evaluations demonstrated that fine-tuned GPT-4o models trained on lower-resolution images achieved up to 7.5% higher accuracy when classifying higher-resolution images. Conversely, models trained on high-resolution images struggled with lower-resolution inputs. Furthermore, cross-plant evaluations indicated that fine-tuned GPT-4o models struggled to generalize disease detection across different plant species, highlighting the necessity of specialized fine-tuning for each crop due to distinct morphological differences in leaf structures.

In the following discussion, each phase of the study will be analyzed in greater detail to extract additional insights and implications for future research.

A. *Evaluation of GPT-4o and ResNet-50 Fine-Tuned Model Performance*

During the initial phase of our study, the ResNet-50 and GPT-4o models were fine-tuned using the same training and validation sets, consisting of plant leaf images annotated with corresponding disease labels. To optimize the hyperparameters for fine-tuning, Bayesian optimization was applied to the ResNet-50 model over 30 trials, exploring a range of hyperparameters and incorporating techniques such as early stopping and pruning. Following fine-tuning at different image resolutions, both models were evaluated on the same test sets to predict plant leaf diseases.

The GPT-4o model consistently outperformed the ResNet-50 model in this task. Specifically, it achieved a 1.24% higher accuracy in detecting diseases in 256px apple leaf images and a 4.37% improvement in corn leaf images. The performance gap widened at lower resolutions, with GPT-4o surpassing ResNet-50 by 3.13% on 150px apple leaves and 8.12% on corn leaves. At 100px resolution, the differences were even more pronounced, with GPT-4o achieving 10.63% and 12.5% higher accuracy on apple and corn leaves, respectively. This suggests that, despite being fine-tuned on lower-resolution images, ResNet-50 is less effective than GPT-4o in this classification task. It is important to note, however, that these models differ significantly in architecture and training scale. ResNet-50 is pre-trained on 25.6 million parameters, whereas GPT-4o is trained on billions of parameters, which may contribute to its superior performance. Architecturally, ResNet-50 is a CNN designed for image recognition, utilizing residual connections to facilitate deeper networks without losing performance. In contrast, GPT-4o is a transformer-based model optimized for natural language processing, leveraging attention mechanisms and a vast number of parameters to capture intricate relationships in large datasets.

However, the higher performance of GPT-4o comes at a substantial cost. Fine-tuning GPT-4o is up to 752.09 times more expensive than fine-tuning ResNet-50, and its fine-tuning process is up to 7.69 times slower. A similar cost disparity is observed in inference, where GPT-4o incurs a fixed cost of $0.23 per 160 image samples, irrespective of resolution. In contrast, the prediction cost for ResNet-50 varies between $0.0051 and $0.0079 per 160 image samples. This significant cost difference is particularly relevant in production environments, where a trade-off between accuracy, speed and computational expense may be necessary.

It is imperative to note that both OpenAI and Google Colab charge for their software-as-a-service (SaaS) offerings and computational infrastructure in U.S. dollars. As such, these cost comparisons would remain proportionally consistent even if expressed in a different currency.

Additionally, the deployment and accessibility of these models differ. GPT-4o is a closed-source API-based model that is easy to deploy and fine-tune, whereas ResNet-50 is an open-source model that requires local deployment and fine-tuning on appropriate GPU hardware, demanding more specialized expertise.

To further illustrate the differences between the two models in plant disease classification, a heatmap representation of a confusion matrix was generated based on their predictions across the test sets for each plant species (**Fig. 7**). This figure shows the accuracy achieved by both models where the diagonal elements represent correct predictions. Prediction accuracy is particularly high in the apple test set, where misclassifications were minimal. However, some misclassifications were observed in the corn test set. Notably, the fine-tuned GPT-4o model misclassified four grey leaf spot samples as rust, while the fine-tuned ResNet-50 model exhibited greater difficulty in correctly classifying northern leaf blight, misclassifying 12 of them as grey leaf spot. These findings provide insights into the weaknesses of each model, enabling targeted adjustments to the fine-tuning process, particularly for real-world deployment scenarios.

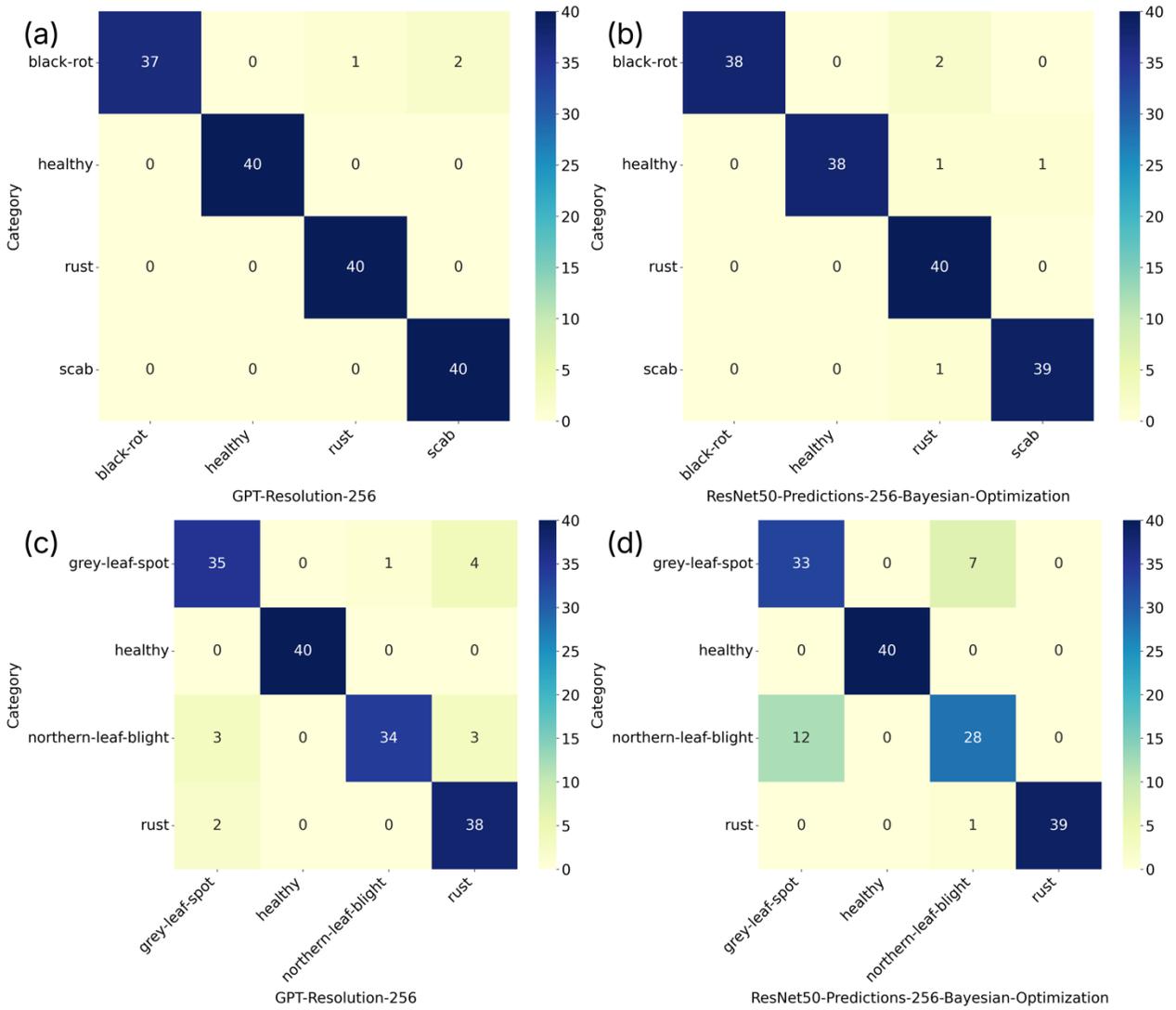

**Fig. 7.** Heatmaps showing prediction distributions of fine-tuned models at a resolution of 256 px: (a) GPT-4o fine-tuned on apple leaf samples, (b) ResNet-50 fine-tuned on apple leaf samples, (c) GPT-4o fine-tuned on corn leaf samples, (d) ResNet-50 fine-tuned on corn leaf samples.

*B. Comparison of Fine-Tuning Hyperparameters*

Sub-sections IV.A and IV.C introduced two distinct methodologies for fine-tuning the GPT-4o model for plant disease detection and classification task. The first approach leveraged hyperparameters obtained through Bayesian optimization of the ResNet-50 model, while the second relied on the default hyperparameters provided by the OpenAI platform.

A comparison of Tables I and III, with additional reference to Tables II and VII.A1, reveals that the latter approach yielded an accuracy that was 2.49% lower than the Bayesian optimization-based approach. At first glance, this finding suggests that Bayesian optimization-derived hyperparameters can be blindly trusted for fine-tuning GPT-4o, given their demonstrated effectiveness.

However, a more skeptical analysis highlights the cost implications of each method. Fine-tuning with default hyperparameters was 3.63 times more cost-effective than using Bayesian-optimized hyperparameters for images of the same resolution. This cost disparity is primarily attributed to the difference in the number of training epochs, 10 epochs for the Bayesian-optimized approach versus 3 epochs for the default hyperparameters. Therefore, a key consideration emerges: whether the 2.49% increase in accuracy justifies the 3.63-fold higher fine-tuning cost, which might be influenced by the specific needs and acceptable tradeoff of target applications.

*C. Identifying Limitations in Zero-Shot Classification*

Comparative analyses using text-based classification demonstrate that LLMs exhibit high accuracy in zero-shot prediction across various domains such as anomaly detection [97], machine translation [98], medical applications [99], and consumer behaviour [100]. However, their performance in image-based classification is notably lower. In our study, GPT-4o achieved accuracy rates of 56.87% for apple leaf disease classification and 68.75% for corn leaf disease classification, despite lacking task-specific training. Given that random chance prediction in this scenario is 25% (1 in 4 possible disease labels per crop), these results indicate a substantial predictive capability. However, in real-world applications where model predictions inform critical decisions—such as selecting appropriate chemical treatments for plant diseases—these accuracy levels remain insufficient. Nevertheless, after fine-tuning with a limited set of labeled samples, the models achieved near-perfect accuracy, underscoring the adaptability of LLMs to domain-specific tasks following targeted training.

*D. Research Findings and Future Directions*

The findings of this study demonstrate that the GPT-4o multimodal LLM is capable of detecting plant leaf diseases, achieving an accuracy of up to 98.12% after fine-tuning. To reach this level of accuracy, a rigorous fine-tuning process was necessary, incorporating a small set of labeled image data. Importantly, this study does not aim to establish whether GPT-4o outperforms other models but rather to confirm that multimodal LLMs can achieve high accuracy in computer vision tasks, even with a limited number of labeled images. This is particularly advantageous in data-constrained scenarios where traditional models struggle due to a scarcity of annotated data.

The superior performance of LLMs in such situations can be attributed to their extensive pretraining phase, which equips them with a robust feature extraction capability. Moreover, LLMs offer additional benefits over CNN-based models, including a superior ability to understand contextual information, their capacity to generalize to previously unseen classes, and the flexibility to incorporate both image and textual data. These attributes make LLMs particularly valuable in dynamic, real-world applications where new conditions or categories may emerge. Furthermore, LLMs' pretraining knowledge allows for a transfer of insights from other domains, providing additional context that enhances prediction accuracy.

A critical factor in leveraging the full potential of multimodal LLMs is optimal prompting. Effective prompt engineering can significantly impact the performance of LLMs, guiding them to interpret the task accurately and align their responses with the required output. In this study, the use of carefully crafted prompts helped the GPT-4o model focus on relevant features and achieve higher accuracy with fewer labeled examples.

However, these advancements are not without limitations. The higher computational cost associated with fine-tuning and inference could be a significant barrier for projects with limited resources. Nonetheless, the versatility of LLMs, enabling not only disease detection but also interactive dialogue and personalized recommendations within the same fine-tuned model adds considerable value to their application. Despite these benefits, certain practical challenges may hinder the widespread adoption of LLMs.

One significant issue encountered during fine-tuning was the exclusion of image samples due to OpenAI's image validation mechanism. This automated system incorrectly flagged certain samples, mistakenly identifying human faces or captchas within the dataset. As observed during progressive fine-tuning (Sections IV.A and IV.C), the set of excluded images differed across different fine-tuning phases, indicating inconsistencies in the validation process. While OpenAI has made substantial progress in this field, such inaccuracies could discourage researchers and businesses from integrating these models into production environments. In practical applications where each data sample is crucial, the exclusion of valid training samples due to algorithmic deficiencies may result in suboptimal model performance.

Another notable finding of this study is the variation in model accuracy when detecting diseases across different plant species. This outcome is expected, given that leaf structures and disease spread patterns differ across plant types. However, it highlights an important avenue for future research. To improve generalizability, future studies should explore fine-tuning LLMs across a broader range of plant species and disease categories. By incorporating diverse datasets, models could be trained to recognize common patterns across various crops, enhancing their robustness in real-world agricultural applications.

Furthermore, future research could investigate hybrid approaches that combine LLMs with specialized vision models to leverage the strengths of both architectures. Exploring self-supervised learning techniques and active learning strategies may also help mitigate data scarcity issues by enabling models to learn from limited labeled data more efficiently. Moreover, the recent surge of vision language models (VLMs) for classification problem such as object detection and 3D object detection could take such approach to next level of AI advancement [101, 102, 103]. Addressing these challenges will be critical in ensuring that LLM-based plant disease detection systems become viable tools for agricultural decision-making and precision farming.

VI. CONCLUSION

The growing threat of plant diseases poses a serious challenge to global food security, sustainable agriculture, and economic stability. Traditional neural network-based approaches often struggle with scalability, accuracy, and adaptability, particularly when labeled data is limited. This study demonstrates the potential of GPT-4o as an effective solution for plant disease classification, showcasing its ability to achieve high accuracy through fine-tuning with minimal labeled data. Specifically, when fine-tuned with few-shot learning, GPT-4o achieved near-zero training loss and high validation accuracy up to 98.12% for apple leaves, outperforming the ResNet-50 baseline, which, although faster and far less costly to fine-tune, exhibited higher losses and lower accuracy, especially at reduced image resolutions. The study also underscores the effectiveness of Bayesian optimization in enhancing fine-tuning outcomes, further improving classification accuracy.

Despite the higher computational costs associated with GPT-4o, its versatility and strong feature extraction capabilities position it as a powerful tool for plant disease detection. The adaptability observed in few-shot learning scenarios reinforces its potential to bridge the gap between conventional computer vision models and advanced multimodal AI solutions.

Additionally, the findings contribute valuable insights into optimizing model performance while maintaining cost efficiency, paving the way for more accessible and scalable AI-driven agricultural solutions. With continued advancements in AI-driven plant disease detection, the integration of such models has the potential to revolutionize precision agriculture, enhancing crop health monitoring and supporting more sustainable farming practices.

**Funding and Acknowledgement:** This work was supported by the National Science Foundation and the United States Department of Agriculture, National Institute of Food and Agriculture through the ``Artificial Intelligence (AI) Institute for Agriculture" Program under Award AWD003473, and AWD004595, Accession Number 1029004, "Robotic Blossom Thinning with Soft Manipulators". The publication of the article in OA mode was financially supported by HEAL-Link.

## VII. Appendix

TABLE A1
FEW-SHOT IMAGE CLASSIFICATION PERFORMANCE
OF PROGRESSIVELY FINE-TUNED GPT-4O MODELS ON THE SAME TEST SET

| Plant | Prediction Column | Fine-Tuned Model | Accuracy | Precision | Recall | F1 | Prediction Duration (e.g., Seconds) | Prediction Cost (e.g., USD) |
|---|---|---|---|---|---|---|---|---|
| Apple | Phase-1-Resolution-256 | B9ojfjfp | 0.9125 | 0.9165 | 0.9125 | 0.9123 | 605.39 | 0.23 |
| | Phase-2-Resolution-256 | B9rgDgVU | 0.9313 | 0.9357 | 0.9312 | 0.9317 | 631.94 | 0.23 |
| | Phase-3-Resolution-256 | B9wnegyN | 0.9563 | 0.9559 | 0.9562 | 0.956 | 684.42 | 0.23 |
| | Phase-4-Resolution-256 | B9zes8zI | 0.9313 | 0.9361 | 0.9312 | 0.9307 | 605.37 | 0.23 |
| | Phase-1-Resolution-150 | BA0XO3aS | 0.8625 | 0.8698 | 0.8625 | 0.8617 | 602.03 | 0.23 |
| | Phase-2-Resolution-150 | BA1jLY4h | 0.9375 | 0.9372 | 0.9375 | 0.9368 | 592.58 | 0.23 |
| | Phase-3-Resolution-150 | BA6q0cyK | 0.9062 | 0.9207 | 0.9062 | 0.9045 | 594.41 | 0.23 |
| | Phase-4-Resolution-150 | BA7oD2kT | 0.9437 | 0.9473 | 0.9437 | 0.9426 | 639.47 | 0.23 |
| | Phase-1-Resolution-100 | BACKwp0N | 0.8438 | 0.8479 | 0.8438 | 0.8411 | 630.29 | 0.23 |
| | Phase-2-Resolution-100 | BAEfIBXZ | 0.7812 | 0.8471 | 0.7812 | 0.772 | 622.34 | 0.23 |
| | Phase-3-Resolution-100 | BAJW3bri | 0.8812 | 0.8949 | 0.8812 | 0.881 | 646.42 | 0.23 |
| | Phase-4-Resolution-100 | BAMVk7Cm | 0.8812 | 0.8836 | 0.8812 | 0.8818 | 626.71 | 0.23 |
| Corn | Phase-1-Resolution-256 | BAWXbcjw | 0.85 | 0.86 | 0.85 | 0.8418 | 624.17 | 0.23 |
| | Phase-2-Resolution-256 | BAXxy46P | 0.8875 | 0.9014 | 0.8875 | 0.8868 | 622.24 | 0.23 |
| | Phase-3-Resolution-256 | BAZKgwpN | 0.9187 | 0.927 | 0.9187 | 0.9178 | 578.94 | 0.23 |
| | Phase-4-Resolution-256 | BAaQ3wfm | 0.925 | 0.93 | 0.925 | 0.9243 | 612.01 | 0.23 |
| | Phase-1-Resolution-150 | BAc2StJe | 0.8688 | 0.8797 | 0.8688 | 0.8627 | 643.66 | 0.23 |
| | Phase-2-Resolution-150 | BB8YhNDx | 0.8938 | 0.898 | 0.8938 | 0.8919 | 592.68 | 0.23 |
| | Phase-3-Resolution-150 | BB9ux708 | 0.9062 | 0.909 | 0.9062 | 0.905 | 522.46 | 0.23 |
| | Phase-4-Resolution-150 | BBAo9aAA | 0.9062 | 0.9083 | 0.9062 | 0.9058 | 650.75 | 0.23 |
| | Phase-1-Resolution-100 | BBBljn0v | 0.8562 | 0.8593 | 0.8562 | 0.8554 | 643.31 | 0.23 |
| | Phase-2-Resolution-100 | BBD8rd0v | 0.8375 | 0.8448 | 0.8375 | 0.8394 | 604.96 | 0.23 |
| | Phase-3-Resolution-100 | BBDySt2J | 0.8688 | 0.8739 | 0.8688 | 0.8705 | 657.02 | 0.23 |
| | Phase-4-Resolution-100 | BBF7QPJj | 0.8812 | 0.8841 | 0.8812 | 0.88 | 621.41 | 0.23 |

Under Review in IEEE TransactionsUnder Review in IEEE Transactions

**References**

[1] Mihrete, T. B., & Mihretu, F. B. (2025). Crop Diversification for Ensuring Sustainable Agriculture, Risk Management and Food Security. *Global Challenges*, 2400267.

[2] Ratnadass, A., Fernandes, P., Avelino, J., & Habib, R. (2012). Plant species diversity for sustainable management of crop pests and diseases in agroecosystems: a review. Agronomy for sustainable development, 32, 273-303.

[3] Abd-Elsalam, K. A., Hassan, R. K., Ahmed, F. K., & Abdelkhalek, T. E. (2024). Plant Health Check: Emerging Methods for Disease Detection. *Plant Quarantine Challenges under Climate Change Anxiety*, 79-124.

[4] Ngugi, H. N., Ezugwu, A. E., Akinyelu, A. A., & Abualigah, L. (2024). Revolutionizing crop disease detection with computational deep learning: a comprehensive review. *Environmental Monitoring and Assessment*, *196*(3), 302.

[5] Upadhyay, A., Chandel, N. S., Singh, K. P., Chakraborty, S. K., Nandede, B. M., Kumar, M., ... & Elbeltagi, A. (2025). Deep learning and computer vision in plant disease detection: a comprehensive review of techniques, models, and trends in precision agriculture. Artificial Intelligence Review, 58(3), 1-64.

[6] Getahun, S., Kefale, H., & Gelaye, Y. (2024). Application of precision agriculture technologies for sustainable crop production and environmental sustainability: A systematic review. *The Scientific World Journal*, *2024*(1), 2126734.

[7] Ruby, E. K., Amirthayogam, G., Sasi, G., Chitra, T., Choubey, A., & Gopalakrishnan, S. (2024). Advanced image processing techniques for automated detection of healthy and infected leaves in agricultural systems. *Mesopotamian Journal of Computer Science*, *2024*, 44-52.

[8] Negi, P., & Anand, S. (2024). Plant disease detection, diagnosis, and management: Recent advances and future perspectives. *Artificial Intelligence and Smart Agriculture: Technology and Applications*, 413-436.

[9] Calabrese, F., Montero-Fernandez, M. A., Kern, I., Pezzuto, F., Lunardi, F., Hofman, P., ... & Galateau-Salle, F. (2024). The role of pathologists in the diagnosis of occupational lung diseases: an expert opinion of the European Society of Pathology Pulmonary Pathology Working Group. *Virchows Archiv*, *485*(2), 173-195.

[10] Karunaweera, N. D., & Dewasurendra, R. (2024). Atypical Leishmania donovani Infections in Sri Lanka: Challenges for Control and Elimination. In *Challenges and Solutions Against Visceral Leishmaniasis* (pp. 163-181). Singapore: Springer Nature Singapore.

[11] Kupa, E., Adanma, U. M., Ogunbiyi, E. O., & Solomon, N. O. (2024). Geologic considerations in agrochemical use: impact assessment and guidelines for environmentally safe farming. *World Journal of Advanced Research and Reviews*, *22*, 1761-1771.

[12] Zhou, W., Li, M., & Achal, V. (2024). A comprehensive review on environmental and human health impacts of chemical pesticide usage. *Emerging Contaminants*, 100410.

[13] MARKOV, T. (2024). EXTERNAL AND INTERNAL THREATS TO FOOD SECURITY. *Scientific Papers Series Management, Economic Engineering in Agriculture & Rural Development*, *24*(2).

[14] Mihrete, T. B., & Mihretu, F. B. (2025). Crop Diversification for Ensuring Sustainable Agriculture, Risk Management and Food Security. *Global Challenges*, 2400267.

[15] Gülmez, B. (2024). Advancements in rice disease detection through convolutional neural networks: A comprehensive review. *Heliyon*.

[16] Abd-Elsalam, K. A., Hassan, R. K., Ahmed, F. K., & Abdelkhalek, T. E. (2024). Plant Health Check: Emerging Methods for Disease Detection. *Plant Quarantine Challenges under Climate Change Anxiety*, 79-124.

[17] Frem, M., Petrontino, A., Fucilli, V., De Lucia, B., Tria, E., Campobasso, A. A., ... & Bozzo, F. (2024). Consumers' Perceptions for an Outdoor Ornamental Plant: Exploring the Influence of Novel Plant Diseases Diagnostics and Sustainable Nurseries Cultivation Management. *Horticulturae*, *10*(5), 501.

[18] Varzakas, T., & Antoniadou, M. (2024). A holistic approach for ethics and sustainability in the food chain: the gateway to oral and systemic health. *Foods*, *13*(8), 1224.

[19] Sajitha, P., Andrushia, A. D., Anand, N., & Naser, M. Z. (2024). A review on machine learning and deep learning image-based plant disease classification for industrial farming systems. *Journal of Industrial Information Integration*, *38*, 100572.

[20] Sabrol, H., & Kumar, S. (2015). Recent studies of image and soft computing techniques for plant disease recognition and classification. *International Journal of Computer Applications*, *126*(1).

[21] Vishnu, S., & Ranjith Ram, A. (2015). Plant disease detection using leaf pattern: A review. *International Journal of Innovative Science, Engineering & Technology*, *2*(6), 774-780.

[22] Martinelli, F., Scalenghe, R., Davino, S., Panno, S., Scuderi, G., Ruisi, P., ... & Dandekar, A. M. (2015). Advanced methods of plant disease detection. A review. *Agronomy for sustainable development*, *35*, 1-25.

[23] Fang, Y., & Ramasamy, R. P. (2015). Current and prospective methods for plant disease detection. *Biosensors*, *5*(3), 537-561.

[24] Wang, H., & Li, H. (2015). Classification recognition of impurities in seed cotton based on local binary pattern and gray level co-occurrence matrix. *Transactions of the Chinese Society of Agricultural Engineering*, *31*(3), 236-241.

[25] Pujari, J. D., Yakkundimath, R., & Byadgi, A. S. (2013). Classification of fungal disease symptoms affected on cereals using color texture features. *International Journal of Signal Processing, Image Processing and Pattern Recognition*, *6*(6), 321-330.

[26] Bankar, S., Dube, A., Kadam, P., & Deokule, S. (2014). Plant disease detection techniques using canny edge detection & color histogram in image processing. *Int. J. Comput. Sci. Inf. Technol.*, *5*(2), 1165-1168.

[27] Revathi, P., & Hemalatha, M. (2012, December). Classification of cotton leaf spot diseases using image processing edge detection techniques. In *2012 International conference on emerging trends in science, engineering and technology (INCOSET)* (pp. 169-173). IEEE.

[28] Weizheng, S., Yachun, W., Zhanliang, C., & Hongda, W. (2008, December). Grading method of leaf spot disease based on image processing. In *2008 international conference on computer science and software engineering* (Vol. 6, pp. 491-494). IEEE.

[29] Khairnar, K., & Dagade, R. (2014). Disease detection and diagnosis on plant using image processing—a review. *International Journal of Computer Applications*, *108*(13), 36-38.

[30] Sankaran, S., Mishra, A., Ehsani, R., & Davis, C. (2010). A review of advanced techniques for detecting plant diseases. *Computers and electronics in agriculture*, *72*(1), 1-13.

[31] Arnal Barbedo, J. G. (2013). Digital image processing techniques for detecting, quantifying and classifying plant diseases. *SpringerPlus*, *2*(1), 660.

[32] Chaerle, L., Hagenbeek, D., De Bruyne, E., Valcke, R., & Van Der Straeten, D. (2004). Thermal and chlorophyll-fluorescence imaging distinguish plant-pathogen interactions at an early stage. *Plant and Cell Physiology*, *45*(7), 887-896.

[33] Moshou, D., Bravo, C., Oberti, R., West, J., Bodria, L., McCartney, A., & Ramon, H. (2005). Plant disease detection based on data fusion of hyper-spectral and multi-spectral fluorescence imaging using Kohonen maps. *Real-Time Imaging*, *11*(2), 75-83.

[34] Ishimwe, R., Abutaleb, K., & Ahmed, F. (2014). Applications of thermal imaging in agriculture—A review. *Advances in remote Sensing*, *3*(3), 128-140.

[35] Yoon, S. C., & Thai, C. N. (2010). Stereo spectral imaging system for plant health characterization. In *Technological developments in networking, education and automation* (pp. 181-186). Dordrecht: Springer Netherlands.

[36] Yang, C., Fernandez, C. J., & Everitt, J. H. (2009). Comparison of airborne multispectral and hyperspectral imagery for mapping cotton root rot. In *2009 Reno, Nevada, June 21-June 24, 2009* (p. 1). American Society of Agricultural and Biological Engineers.

[37] Terentev, A., Dolzhenko, V., Fedotov, A., & Eremenko, D. (2022). Current state of hyperspectral remote sensing for early plant disease detection: A review. *Sensors*, *22*(3), 757.

[38] Demilie, W. B. (2024). Plant disease detection and classification techniques: a comparative study of the performances. *Journal of Big Data*, *11*(1), 5.

[39] Ojo, M. O., & Zahid, A. (2023). Improving deep learning classifiers performance via preprocessing and class imbalance approaches in a plant disease detection pipeline. *Agronomy*, *13*(3), 887.

[40] Ngugi, H. N., Akinyelu, A. A., & Ezugwu, A. E. (2024). Machine Learning and Deep Learning for Crop Disease Diagnosis: Performance Analysis and Review. *Agronomy*, *14*(12), 3001.

[41] Li, L., Zhang, S., & Wang, B. (2021). Plant disease detection and classification by deep learning—a review. *IEEE Access*, *9*, 56683-56698.

[42] Kolli, J., Vamsi, D. M., & Manikandan, V. M. (2021, November). Plant disease detection using convolutional neural network. In *2021 IEEE Bombay Section Signature Conference (IBSSC)* (pp. 1-6). IEEE.

[43] Chen, H. C., Widodo, A. M., Wisnujati, A., Rahaman, M., Lin, J. C. W., Chen, L., & Weng, C. E. (2022). AlexNet convolutional neural network for disease detection and classification of tomato leaf. *Electronics*, *11*(6), 951.

Under Review in IEEE Transactions[44] Matin, M. M. H., Khatun, A., Moazzam, M. G., & Uddin, M. S. (2020). An efficient disease detection technique of rice leaf using AlexNet. *Journal of Computer and Communications*, *8*(12), 49-57.

[45] Li, Z., Li, C., Deng, L., Fan, Y., Xiao, X., Ma, H., ... & Zhu, L. (2022). Improved AlexNet with Inception-V4 for Plant Disease Diagnosis. *Computational intelligence and neuroscience*, *2022*(1), 5862600.

[46] Alatawi, A. A., Alomani, S. M., Alhawiti, N. I., & Ayaz, M. (2022). Plant disease detection using AI based VGG-16 model. *International Journal of Advanced Computer Science and Applications*, *13*(4).

[47] Paymode, A. S., & Malode, V. B. (2022). Transfer learning for multi-crop leaf disease image classification using convolutional neural network VGG. *Artificial Intelligence in Agriculture*, *6*, 23-33.

[48] Kumar, A., Razi, R., Singh, A., & Das, H. (2020, June). Res-vgg: A novel model for plant disease detection by fusing vgg16 and resnet models. In *International conference on machine learning, image processing, network security and data sciences* (pp. 383-400). Singapore: Springer Singapore.

[49] Jiangqing, W., Xing, J., Haifang, M., Jun, T., & Chang, L. (2022). Plant disease detection based on lightweight VGG. *Journal of Chinese Agricultural Mechanization*, *43*(4), 25.

[50] Archana, U., Khan, A., Sudarshanam, A., Sathya, C., Koshariya, A. K., & Krishnamoorthy, R. (2023, April). Plant disease detection using resnet. In *2023 International Conference on Inventive Computation Technologies (ICICT)* (pp. 614-618). IEEE.

[51] Kumar, V., Arora, H., & Sisodia, J. (2020, July). Resnet-based approach for detection and classification of plant leaf diseases. In *2020 international conference on electronics and sustainable communication systems (ICESC)* (pp. 495-502). IEEE.

[52] Reddy, S. R., Varma, G. S., & Davuluri, R. L. (2023). Resnet-based modified red deer optimization with DLCNN classifier for plant disease identification and classification. *Computers and Electrical Engineering*, *105*, 108492.

[53] Li, X., & Rai, L. (2020, November). Apple leaf disease identification and classification using resnet models. In *2020 IEEE 3rd International Conference on Electronic Information and Communication Technology (ICEICT)* (pp. 738-742). IEEE.

[54] Kalaivani, S., Tharini, C., Viswa, T. S., Sara, K. F., & Abinaya, S. T. (2025). ResNet-based classification for leaf disease detection. *Journal of The Institution of Engineers (India): Series B*, *106*(1), 1-14.

[55] Joseph, D. S., Pawar, P. M., & Pramanik, R. (2023). Intelligent plant disease diagnosis using convolutional neural network: a review. *Multimedia Tools and Applications*, *82*(14), 21415-21481.

[56] Lu, J., Tan, L., & Jiang, H. (2021). Review on convolutional neural network (CNN) applied to plant leaf disease classification. *Agriculture*, *11*(8), 707.

[57] Abbas, A., Jain, S., Gour, M., & Vankudothu, S. (2021). Tomato plant disease detection using transfer learning with C-GAN synthetic images. *Computers and electronics in agriculture*, *187*, 106279.

[58] Dong, J., Fuentes, A., Zhou, H., Jeong, Y., Yoon, S., & Park, D. S. (2024). The impact of fine-tuning paradigms on unknown plant diseases recognition. *Scientific Reports*, *14*(1), 17900.

[59] Zhang, S., & Zhang, C. (2023). Modified U-Net for plant diseased leaf image segmentation. *Computers and Electronics in Agriculture*, *204*, 107511.

[60] Liu, W., Yu, L., & Luo, J. (2022). A hybrid attention-enhanced DenseNet neural network model based on improved U-Net for rice leaf disease identification. *Frontiers in Plant Science*, *13*, 922809.

[61] Bondre, S., & Patil, D. (2024). Crop disease identification segmentation algorithm based on Mask-RCNN. *Agronomy Journal*, *116*(3), 1088-1098.

[62] Rehman, Z. U., Khan, M. A., Ahmed, F., Damaševičius, R., Naqvi, S. R., Nisar, W., & Javed, K. (2021). Recognizing apple leaf diseases using a novel parallel real-time processing framework based on MASK RCNN and transfer learning: An application for smart agriculture. *IET Image Processing*, *15*(10), 2157-2168.

[63] Kumar, M., Chandel, N. S., Singh, D., & Rajput, L. S. (2023). Soybean disease detection and segmentation based on Mask-RCNN algorithm. *J Exp Agric Int*, *45*(5), 63-72.

[64] Pallapothu, T., Singh, M., Sinha, R., Nangia, H., & Udawant, P. (2022, May). Cotton leaf disease detection using mask RCNN. In *AIP Conference Proceedings* (Vol. 2393, No. 1). AIP Publishing.

[65] Kumar, D., & Kukreja, V. (2022, March). Image-based wheat mosaic virus detection with Mask-RCNN model. In *2022 international conference on decision aid sciences and applications (DASA)* (pp. 178-182). IEEE.

[66] Gomaa, A. A., & Abd El-Latif, Y. M. (2021). Early prediction of plant diseases using CNN and GANs. *International Journal of Advanced Computer Science and Applications*, *12*(5).

[67] Singh, A. K., Rao, A., Chattopadhyay, P., Maurya, R., & Singh, L. (2024). Effective plant disease diagnosis using Vision Transformer trained with leafy-generative adversarial network-generated images. *Expert Systems with Applications*, *254*, 124387.

[68] Wu, Y., & Xu, L. (2021). Image generation of tomato leaf disease identification based on adversarial-VAE. *Agriculture*, *11*(10), 981.

[69] Akkem, Y., Biswas, S. K., & Varanasi, A. (2024). A comprehensive review of synthetic data generation in smart farming by using variational autoencoder and generative adversarial network. *Engineering Applications of Artificial Intelligence*, *131*, 107881.

[70] Mekha, P., & Teeyasuksaet, N. (2021, March). Image classification of rice leaf diseases using random forest algorithm. In *2021 joint international conference on digital arts, media and technology with ECTI northern section conference on electrical, electronics, computer and telecommunication engineering* (pp. 165-169). IEEE.

[71] Sahu, S. K., & Pandey, M. (2023). An optimal hybrid multiclass SVM for plant leaf disease detection using spatial Fuzzy C-Means model. *Expert systems with applications*, *214*, 118989.

[72] Wei, T., Chen, Z., & Yu, X. (2024, December). Snap and diagnose: An advanced multimodal retrieval system for identifying plant diseases in the wild. In *Proceedings of the 6th ACM International Conference on Multimedia in Asia* (pp. 1-3).

[73] Liu, J., & Wang, X. (2024). A multimodal framework for pepper diseases and pests detection. *Scientific Reports*, *14*(1), 28973.

[74] Wei, T., Chen, Z., Huang, Z., & Yu, X. (2024, October). Benchmarking in-the-wild multimodal disease recognition and a versatile baseline. In *Proceedings of the 32nd ACM International Conference on Multimedia* (pp. 1593-1601).

[75] Lu, Y., Lu, X., Zheng, L., Sun, M., Chen, S., Chen, B., ... & Lv, C. (2024). Application of multimodal transformer model in intelligent agricultural disease detection and question-answering systems. *Plants*, *13*(7), 972.

[76] Yan, R., An, P., Meng, X., Li, Y., Li, D., Xu, F., & Dang, D. (2025). A knowledge graph for crop diseases and pests in China. *Scientific Data*, *12*(1), 222.

[77] Yang, W., Yang, S., Wang, G., Liu, Y., Lu, J., & Yuan, W. (2023). Knowledge graph construction and representation method for potato diseases and pests. *Agronomy*, *14*(1), 90.

[78] "GitHub - Applied-AI-Research-Lab/Multimodal-Large-Language-Models-in-Agriculture: Multimodal Large Language Models in Agriculture." Accessed: Mar. 21, 2025. [Online]. Available: https://github.com/Applied-AI-Research-Lab/Multimodal-Large-Language-Models-in-Agriculture

[79] "GitHub - spMohanty/PlantVillage-Dataset: Dataset of diseased plant leaf images and corresponding labels." Accessed: Mar. 21, 2025. [Online]. Available: https://github.com/spMohanty/PlantVillage-Dataset

[80] S. P. Mohanty, D. P. Hughes, and M. Salathé, "Using deep learning for image-based plant disease detection," Front Plant Sci, vol. 7, no. September, Sep. 2016, doi: 10.3389/FPLS.2016.01419.

[81] V. V. Srinidhi, A. Sahay, and K. Deeba, "Plant Pathology Disease Detection in Apple Leaves Using Deep Convolutional Neural Networks : Apple Leaves Disease Detection using EfficientNet and DenseNet," Proceedings - 5th International Conference on Computing Methodologies and Communication, ICCMC 2021, pp. 1119–1127, Apr. 2021, doi: 10.1109/ICCMC51019.2021.9418268.

[82] Q. Wu et al., "A classification method for soybean leaf diseases based on an improved ConvNeXt model," Scientific Reports 2023 13:1, vol. 13, no. 1, pp. 1–11, Nov. 2023, doi: 10.1038/s41598-023-46492-3.

[83] U. Barman et al., "ViT-SmartAgri: Vision Transformer and Smartphone-Based Plant Disease Detection for Smart Agriculture," Agronomy 2024, Vol. 14, Page 327, vol. 14, no. 2, p. 327, Feb. 2024, doi: 10.3390/AGRONOMY14020327.

[84] S. R. G. Reddy, G. P. S. Varma, and R. L. Davuluri, "Resnet-based modified red deer optimization with DLCNN classifier for plant disease identification and classification," Computers and Electrical Engineering, vol. 105, p. 108492, Jan. 2023, doi: 10.1016/J.COMPELECENG.2022.108492.

[85] V. S. Desanamukula, T. D. Teja, and P. Rajitha, "An In-Depth Exploration of ResNet-50 and Transfer Learning in Plant Disease Diagnosis," 7th International Conference on Inventive Computation Technologies, ICICT 2024, pp. 614–621, 2024, doi: 10.1109/ICICT60155.2024.10544802.

[86] "Introducing vision to the fine-tuning API | OpenAI." Accessed: Dec. 27, 2024. [Online]. Available: https://openai.com/index/introducing-vision-to-the-fine-tuning-api/

[87] K. I. Roumeliotis, N. D. Tselikas, and D. K. Nasiopoulos, "LLMs in e-commerce: A comparative analysis of GPT and LLaMA models in product review evaluation," Natural Language Processing Journal, vol. 6, p. 100056, Mar. 2024, doi: 10.1016/J.NLP.2024.100056.
23 | P a g e

[88] K. Zhang, F. Zhou, L. Wu, N. Xie, and Z. He, "Semantic understanding and prompt engineering for large-scale traffic data imputation," Information Fusion, vol. 102, p. 102038, Feb. 2024, doi: 10.1016/J.INFFUS.2023.102038.
[89] Anthropic PBC, "Automatically generate first draft prompt templates - Anthropic." Accessed: Nov. 30, 2024. [Online]. Available: https://docs.anthropic.com/en/docs/build-with-claude/prompt-engineering/prompt-generator
[90] N. S. Keskar, J. Nocedal, P. T. P. Tang, D. Mudigere, and M. Smelyanskiy, "On Large-Batch Training for Deep Learning: Generalization Gap and Sharp Minima," 5th International Conference on Learning Representations, ICLR 2017 - Conference Track Proceedings, Sep. 2016, Accessed: Mar. 22, 2025. [Online]. Available: https://arxiv.org/abs/1609.04836v2
[91] D. R. Wilson and T. R. Martinez, "The need for small learning rates on large problems," Proceedings of the International Joint Conference on Neural Networks, vol. 1, pp. 115–119, 2001, doi: 10.1109/IJCNN.2001.939002.
[92] H. Li, J. Li, X. Guan, B. Liang, Y. Lai, and X. Luo, "Research on Overfitting of Deep Learning," Proceedings - 2019 15th International Conference on Computational Intelligence and Security, CIS 2019, pp. 78–81, Dec. 2019, doi: 10.1109/CIS.2019.00025.
[93] R. Shirvalkar and A. S. Remya Ajai, "Enhancing Pneumonia Detection Accuracy Through ResNet-Based Deep Learning Models and Ensemble Techniques: A Study Using Chest X-Ray Images," Lecture Notes in Networks and Systems, vol. 949 LNNS, pp. 25–37, 2024, doi: 10.1007/978-981-97-1313-4_3.
[94] K. Hamed and U. Ozgunalp, "A Comparative Analysis of Pretrained Models for Brain Tumaor Classification and Their Optimization Using Optuna," 2024 Innovations in Intelligent Systems and Applications Conference, ASYU 2024, 2024, doi: 10.1109/ASYU62119.2024.10757117.
[95] M. A. K. Raiaan et al., "A Review on Large Language Models: Architectures, Applications, Taxonomies, Open Issues and Challenges," IEEE Access, vol. 12, pp. 26839–26874, 2024, doi: 10.1109/ACCESS.2024.3365742.
[96] I. Bouacida, B. Farou, L. Djakhdjakha, H. Seridi, and M. Kurulay, "Innovative deep learning approach for cross-crop plant disease detection: A generalized method for identifying unhealthy leaves," Information Processing in Agriculture, vol. 12, no. 1, pp. 54–67, Mar. 2025, doi: 10.1016/J.INPA.2024.03.002.
[97] J. Zhu, S. Cai, F. Deng, and J. Wu, "Do LLMs Understand Visual Anomalies? Uncovering LLM Capabilities in Zero-shot Anomaly Detection," Apr. 2024, doi: 10.1145/3664647.3681190.
[98] B. Ji, X. Duan, Y. Zhang, K. Wu, and M. Zhang, "Zero-shot Prompting for LLM-based Machine Translation Using In-domain Target Sentences," IEEE/ACM Trans Audio Speech Lang Process, 2024, doi: 10.1109/TASLP.2024.3519814.
[99] J. J. D. Leow, H. N. Chua, M. B. Jasser, B. Issa, and R. T. K. Wong, "Comparison of Depression Detection Between LLMs and Zero-Shot Learning Using DAD Dataset," 2025 21st IEEE International Colloquium on Signal Processing & Its Applications (CSPA), pp. 295–300, Feb. 2025, doi: 10.1109/CSPA64953.2025.10933098.
[100] K. I. Roumeliotis, N. D. Tselikas, and D. K. Nasiopoulos, "Think Before You Classify: The Rise of Reasoning Large Language Models for Consumer Complaint Detection and Classification," Electronics 2025, Vol. 14, Page 1070, vol. 14, no. 6, p. 1070, Mar. 2025, doi: 10.3390/ELECTRONICS14061070.
[101] Sapkota, R., & Karkee, M. Object Detection with Multimodal Large Vision-Language Models: An In-depth Review. Authorea TechRxiv. April 25, 2025. DOI: 10.36227/techrxiv.174559706.69198342/v1
[102] R. Sapkota, K. I. Roumeliotis, R. H. Cheppally, M. Flores Calero, and M. Karkee, "A Review of 3D Object Detection with Vision-Language Models," arXiv preprint arXiv:2504.18738, 2025. https://doi.org/10.48550/arXiv.2504.18738
[103] Sapkota, Ranjan and Karkee, Manoj, Object Detection with Multimodal Large Vision-Language Models: An In-Depth Review. Available at SSRN: https://ssrn.com/abstract=5233953 or http://dx.doi.org/10.2139/ssrn.5233953

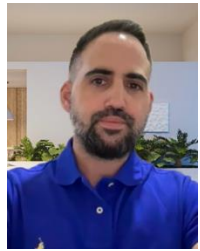

**KONSTANTINOS I. ROUMELIOTIS** is an Adjunct Professor at the Department of Digital Systems and a postdoctoral researcher at the Department of Informatics and Telecommunications, University of Peloponnese. With a strong focus on the intersection of applied artificial intelligence and business, his research explores innovative solutions in deep learning, multimodal large language models, and natural language processing, aiming to drive impactful advancements in real-world business applications.

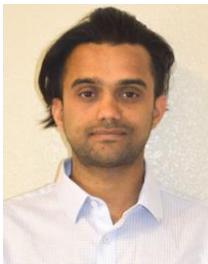

**RANJAN SAPKOTA** is a doctoral student at Cornell University in the Department of Biological and Environmental Engineering, where his research centers on agricultural automation and robotics, with a particular focus on artificial intelligence technologies such as deep learning, large language models, and vision-language models. He received his M.S. in Agricultural and Biosystems Engineering from North Dakota State University in 2022 and earned a B.Tech. in Electrical and Electronics Engineering in 2019.

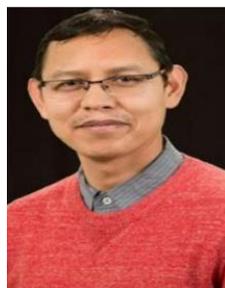

**MANOJ KARKEE** He is currently a Professor at Cornell University, Department of Biological and Environmental Engineering. His research interests include agricultural automation and mechanization programs, with an emphasis on machine vision-based sensing, automation, and robotic technologies for specialty crop production. He received the B.S. degree in computer engineering from Tribhuvan University, in 2002, the M.S. degree in remote sensing and geographic information systems from Asian Institute of Technology, Thailand, and the Ph.D. degree in agricultural engineering and human–computer interaction from Iowa State University, in 2009.

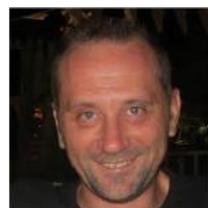

**NIKOLAOS D. TSELIKAS** is a Professor in the Department of Informatics and Telecommunications, at the University of Peloponnese (UoP), in Greece. His research focuses on areas such as software engineering, software performance evaluation, open source software and computer vision applications. He has led several funded research projects related to software engineering methodologies and computer vision technologies and applications. He has authored over 100 publications in international journals, conferences, and book chapters, earning significant recognition for his work. Finally, his book on C programming language is widely adopted as a recommended textbook in numerous universities worldwide.

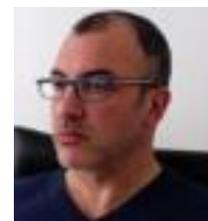

**DIMITRIOS K. NASIOPOULOS** is an Ass. Professor at the Department of Agribusiness and Supply Chain Management of the Agricultural University of Athens. His scientific/research work includes scientific interests such as Management Information Systems, Decision-Making systems, Simulation Modeling, Business Intelligence, Business Informatics, Computational Methods and Management of Supply Chain Operations, LLM and NLP models, etc. Also, he has several research projects including modelling and simulation in business area, web-based programming, databases and web applications. He is editor-in-chief in Springer book series and Chairman of the Organizing Committee of international conferences and internationally renowned journals in the publishing houses Emerald, Elsevier, Springer for a number of years, as well as a member of the editorial board in several scientific journals.